\documentclass[letterpaper,11pt, table, xcdraw]{article}
\usepackage[hyphens]{url}

\usepackage[preprint]{acl}

\usepackage{times}
\usepackage{latexsym}
\usepackage[T1]{fontenc}
\usepackage[utf8]{inputenc}
\usepackage{microtype}
\usepackage{inconsolata}
\usepackage{helvet}
\usepackage{courier}
\usepackage[hyphens]{url}
\usepackage{graphicx}
\usepackage{natbib}
\usepackage{caption}
\usepackage{algorithm}
\usepackage{algorithmic}
\usepackage{amsmath}
\usepackage{amssymb}
\newcommand{\llmname}[1]{{\fontfamily{pcr}\selectfont {#1}}\xspace}
\usepackage{xspace}

\usepackage[utf8]{inputenc}
\usepackage{pifont}
\newcommand{\myding}[1]{{\normalfont\ding{#1}}}
\newenvironment{mydinglist}[1]
{%
  \normalfont
  \begin{dinglist}{#1}
}
{%
  \end{dinglist}
}

\usepackage{longtable}
\usepackage[most]{tcolorbox}
\usepackage{multirow}
\usepackage{xcolor}
\usepackage{graphicx}
\usepackage{subcaption}

\usepackage{newfloat}
\usepackage{listings}

\usepackage[normalem]{ulem}
\useunder{\uline}{\ul}{}

\DeclareCaptionStyle{ruled}{labelfont=normalfont,labelsep=colon,strut=off} 
\floatstyle{ruled}
\newfloat{listing}{tb}{lst}{}
\floatname{listing}{Listing}
\title{Learning to Edit Knowledge via Instruction-based Chain-of-Thought Prompting}

\author{
 \textbf{Jinhu Fu\textsuperscript{1}}\thanks{fjhu@bupt.edu.cn},
 \textbf{Yan Bai\textsuperscript{2}},
 \textbf{Longzhu He\textsuperscript{1}},
 \textbf{Yihang Lou\textsuperscript{3}},
\\
 \textbf{Yanxiao Zhao\textsuperscript{1}},
 \textbf{Li Sun\textsuperscript{1}},
 \textbf{Sen Su\textsuperscript{1,4}}\thanks{Corresponding author.},
\\
 \textsuperscript{1}Beijing University of Posts and Telecommunications,
 \textsuperscript{2}Peking University,
\\
 \textsuperscript{3}Huawei Technologies Ltd.,
 \textsuperscript{4}Chongqing University of Posts and Telecommunications
}

\begin{document}
\maketitle
\begin{abstract}
 Large language models (LLMs) can effectively handle outdated information through knowledge editing. However, current approaches face two key limitations: \textbf{(I)} \textit{\textbf{Poor generalization}}: Most approaches rigidly inject new knowledge without ensuring that the model can use it effectively to solve practical problems. \textbf{(II)} \textit{\textbf{Narrow scope}}: Current methods focus primarily on structured fact triples, overlooking the diverse unstructured forms of factual information (e.g., news, articles) prevalent in real-world contexts. 
 To address these challenges, we propose a new paradigm: teaching LLMs to edit knowledge via \textit{\textbf{Chain of Thoughts (CoTs)}} reasoning (\llmname{CoT2Edit}). We first leverage language model agents for both structured and unstructured edited data to generate CoTs, building high-quality instruction data. 
 The model is then trained to reason over edited knowledge through supervised fine-tuning (SFT) and Group Relative Policy Optimization (GRPO). At inference time, we integrate Retrieval-Augmented Generation (RAG) to dynamically retrieve relevant edited facts for real-time knowledge editing.  Experimental results demonstrate that our method achieves strong generalization across six diverse knowledge editing scenarios with just \textbf{a single round of training} on three open-source language models. The codes are available at \url{https://github.com/FredJDean/CoT2Edit}.
\end{abstract}

\section{Introduction}

The advent of large language models (LLMs) has brought revolutionary advancements to the field of natural language processing \cite{radford2019language, roberts2020much}. However, LLMs may harbor incorrect or outdated knowledge \cite{mitchellfast, de2021editing}, and knowledge editing methods have emerged as a solution to this issue \cite{zhang2024comprehensive, 2025arXiv250415585W}. 
Mainstream approaches to LLM knowledge editing include context-based methods and parameter-modification-based methods \cite{zhang2024comprehensive}. Context-based methods, represented by works such as \cite{qi2024context, zheng2023can, chen-etal-2024-robust}, offer broad applicability but require the design of complex prompt templates and exhibit sub-optimal performance \cite{li2024pmet, yu2024melo, wang2024wise}. In contrast, parameter-modification-based methods have garnered widespread attention due to their more stable performance \cite{li2024pmet, gu2024model,fu2026diagnosingrepairingunsafechannels}. 
Among these, locate-then-edit methods have achieved remarkable progress in recent years \cite{fang2024alphaedit,2025arXiv250205628J}, with representative works including  \llmname{MEMIT} \cite{meng2022mass}, \llmname{AlphaEdit} \cite{fang2024alphaedit}, \llmname{KGMET}\cite{fu-etal-2025-knowledge} and others \cite{li2024pmet, gu2024model}.

\begin{figure*}
  \centering
  \includegraphics[height=7.0cm]{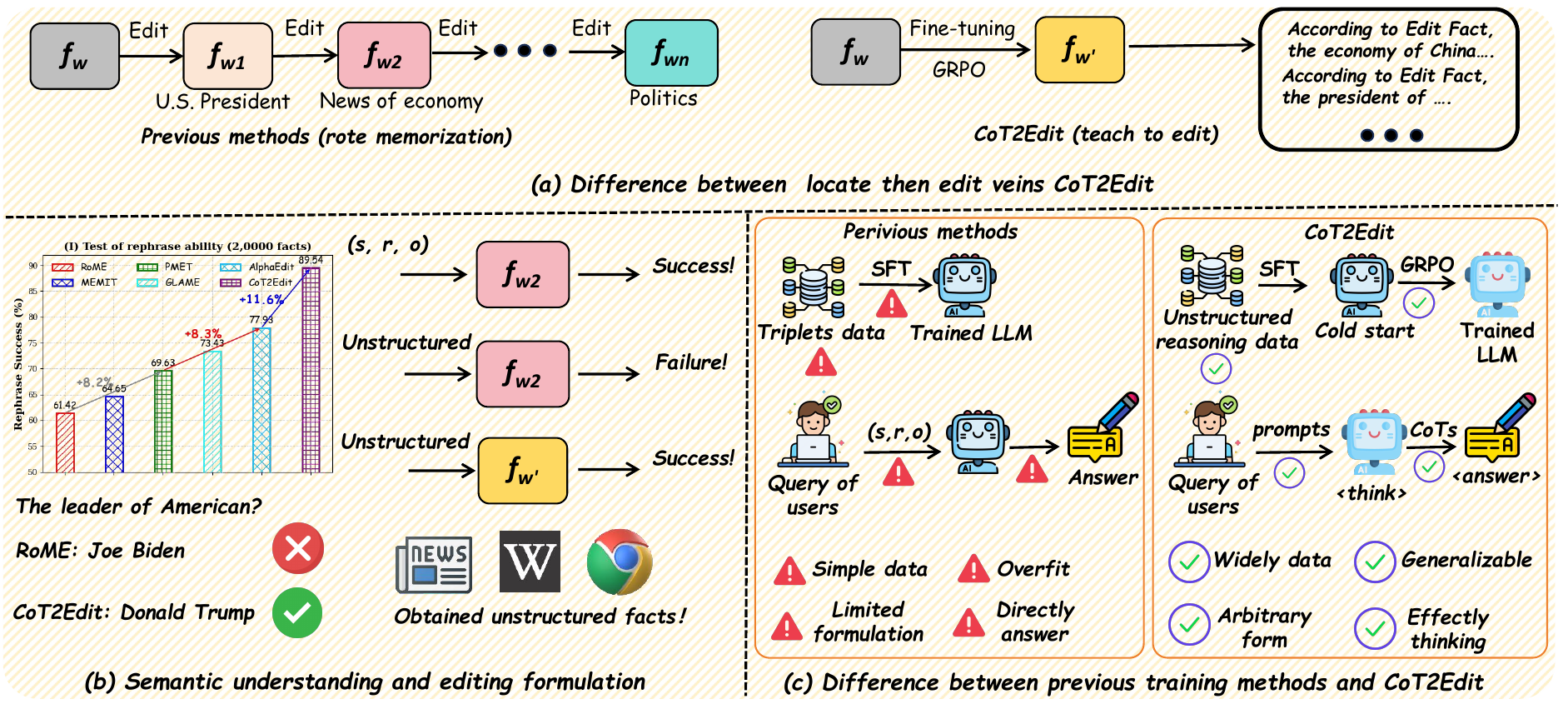}
  \caption{Revealing the deficiencies in existing knowledge editing approaches.}\label{fig:1}
\end{figure*}

However, locate-then-edit approaches lack both \textbf{flexible utilization} and \textbf{semantic understanding} capabilities. \ding{172} As shown in Figure \ref{fig:1} (a), these require direct access to and modification of model parameters, which is incompatible with frozen, production-grade LLMs \cite{jiang-etal-2024-learning}. \ding{173} As illustrated in Figure \ref{fig:1} (b), when editing knowledge at the parameter level, the model may respond correctly to exact queries (e.g., "the president of the U.S.") but fail on semantically equivalent ones (e.g., "the U.S. leader") \cite{EditCoT}. In addition, new knowledge often appears in unstructured forms, such as news or articles, rather than rigid triples, causing it fail to handle such knowledge \cite{AKEW}. 

To address the limitations, some works such as \llmname{LTE} \cite{jiang-etal-2024-learning} and \llmname{EditCoT} \cite{EditCoT} adopt a training and retrieval paradigm. Though effectiveness, these approaches still suffer from several critical limitations:

\ding{224} \textbf{\textit{Poor Availability.}}
\llmname{LTE} \cite{jiang-etal-2024-learning} is trained on simple editing datasets that do not explicitly model reasoning paths. As a result, the model is required to generate correct answers in a single step, which makes it \textbf{prone to hallucinations}. \llmname{EditCoT} \cite{EditCoT} partially addresses this issue by introducing CoTs, but it relies on a complex multi-model pipeline with \textbf{separate LLMs for reasoning and editing}, increasing computational cost while limiting scalability in practical deployment. In addition, they focus mainly on editing simple factual triples, making them difficult to extend to \textbf{more realistic knowledge forms}, such as unstructured or compositional knowledge.

\ding{224} \textbf{\textit{Limited generalization}}. Both \llmname{LTE} and \llmname{EditCoT} depend heavily on large-scale SFT \cite{howard2018universal} over curated editing datasets to learn specific editing patterns. While effective in-distribution, this strategy leads to over-fitting to the training data and results in noticeable performance degradation when scaling to diverse or large-scale editing settings. Moreover, when confronted with out-of-distribution (OOD) editing data, their performance remains limited, highlighting the \textbf{inherent generalization bottleneck} of purely SFT-based editing approaches \cite{kang2025quagmires}.

To address these challenges, we propose \llmname{CoT2Edit}. As shown in Figure \ref{fig:1} (c), we redesign knowledge editing paradigm from the perspectives of data, training and inference. \ding{182} For data construction, \underline{to solve the narrow scope problem}, we cover a broader range of editing scenarios by jointly modeling structured and unstructured edits, using MQUAKE \cite{zhong2023mquake} and MQUAKE-uns \cite{AKEW} to generate CoTs supervision and synthesizing additional instruction data from HotpotQA \cite{hotpotqa} entity relations with LLM agents (e.g. \llmname{ChatGPT} or \llmname{DeepSeek}). \ding{183} For training, we adopt a two-stage SFT and GRPO \cite{grpo} framework, where SFT only provides a cold start, while data evolved GRPO serves as the core optimization, effectively \underline{improving generalization to unseen edits} while preserving the general capabilities of LLMs. \ding{184} For inference, \llmname{CoT2Edit} performs CoTs reasoning over retrieved edited facts, which \underline{mitigates hallucinations from single-step generation} and enhances the ability of edited LLMs to solve complex editing queries.


Experimental results demonstrate that our proposed \llmname{CoT2Edit} excels in three key aspects:
\myding{182} \textbf{Single-Training Generalization:} Unlike previous approaches requiring repeated editing processes, \llmname{CoT2Edit} achieves state-of-the-art performance on six previously unseen editing benchmarks  after just a single training process, demonstrating remarkable out-of-distribution (OOD) generalization capability. \myding{183} \textbf{Cross-Format Robustness}. The proposed paradigm shows consistent excellence in unstructured knowledge editing scenarios. Specifically, our approach achieves 92\% accuracy for unstructured knowledge editing($\uparrow \sim$ 20\% compared with \llmname{IKE} \cite{zheng2023can}). \myding{184} \textbf{Semantic Robustness}. \llmname{CoT2Edit} achieves 89\% in the rephrase metric ($\uparrow \sim$ 8\%) and 93\% in the neighborhood success metric ($\uparrow \sim$ 23\%) when editing \textbf{large scale} facts (20k-30k vs. 2k-3k in \llmname{AlphaEdit} \cite{fang2024alphaedit}). 

Our contributions can be summarized as follows: 
\begin{mydinglist}{43}
    \item We construct a novel editing training dataset that incorporates multi-hop editing tasks to generate CoTs instruction data. Our dataset integrates both structured and unstructured knowledge sources, covering diverse editing scenarios beyond triplet modifications.

    \item We are the first to apply GRPO reinforcement learning to knowledge editing, where a lightweight SFT cold start is followed by data-evolving GRPO optimization, enabling LLMs to effectively generalize to complex reasoning tasks.

    \item Our method achieves superior performance across six standard knowledge editing benchmarks using only a single training round, highlighting its strong out-of-distribution capability and cross-format robustness. 
\end{mydinglist}

\section{Related works}
In this section, we introduce related work on model editing. \textbf{(I) In-context editing methods}. Representative works in this category include \llmname{ICE} \cite{qi2024context}, \llmname{IKE} \cite{zheng2023can}, and so on. This approach is typically designed for black-box scenarios and does not allow for parameter adjustments \cite{zhang2024comprehensive}. \texttt{LTE} \cite{jiang-etal-2024-learning} fine-tunes LLMs using corpora built from editing targets and related knowledge, but it struggles with multi-hop reasoning and complex edits. \texttt{EditCoT} \cite{EditCoT} enhances the confidence of edited knowledge by transplanting chains of thought related to the editing target into the LLM’s context. Nevertheless, \texttt{EditCoT} requires first fine-tuning a model to generate chains of thought for the new knowledge, and then transferring them into the context of another editing model, which is both time-consuming and resource-intensive.  \textbf{(II) Side memory based methods}. The works such as \llmname{MELo} \cite{yu2024melo}, \llmname{WISE} \cite{wang2024wise} and \llmname{GRACE} \cite{hartvigsen2024aging}. They edit external memory to achieve knowledge update. Although effective, as the number of edits increases, they become memory-overhead, making it cumbersome. \textbf{(III) Locate then edit methods}. \llmname{RoME}\cite{meng2022locating} first locates relevant neurons via causal mediation trace \cite{pearl2001direct, vig2020investigating} and modifies the corresponding MLP module. Building upon this, \llmname{MEMIT} \cite{meng2022mass} introduced batch editing, which allows efficient updates to large-scale knowledge. In recent years, approaches such as \llmname{AlphaEdit} \cite{fang2024alphaedit}, \llmname{PMET} \cite{li2024pmet}, and \llmname{GLAME} \cite{zhang2024knowledge} have been proposed, significantly advancing the upper bound of performance for the locate-then-edit paradigm. Although effective, they inherently suffer from the limitations of “rote memorization” when modifying K-V pairs.

\section{Preliminary}
Knowledge editing \cite{mitchellfast} aims to update the base model $f_\theta$ so that it reflects a desired change specified by an edit descriptor $(x_i, y_i)$, while preserving performance on unrelated inputs. Formally, given that the original model predicts $f_{\theta}(x_i)\neq y_i$, the goal is to obtain an updated model $f_{\theta'}$ such that $f_{\theta'}(x_i)=y_i$.

Formally, our objective is to learn a mapping function $f_{\theta'}: \mathcal{E} \times \mathcal{Q} \rightarrow \mathcal{A}$, where $\mathcal{E}$ denotes editable facts, $\mathcal{Q}$ represents possible queries, and $\mathcal{A}$ denotes the answer space. such that for any editing fact $e \in \mathcal{E}$ and question $q \in \mathcal{Q}$, the function gives $a = f_{\theta'}(e,q)$ as the updated response.

However, unlike previous editing paradigms that directly enforce the edited fact during prediction, \texttt{CoT2Edit} introduces an intermediate reasoning process. Specifically, we reformulate the mapping as a two-stage function:

\begin{equation}
\label{eq.2}
f_{\theta'}(e, q) = g_{\theta'}(h_{\theta'}(e, q))
\end{equation}

where $h_{\theta'}(e, q)$ generates an interpretable reasoning chain conditioned on the edit fact $e$ and query $q$, and $g_{\theta'}(\cdot)$ produces the final answer based on this reasoning. 

\begin{figure*}
  \centering
  \includegraphics[height=8.9cm]{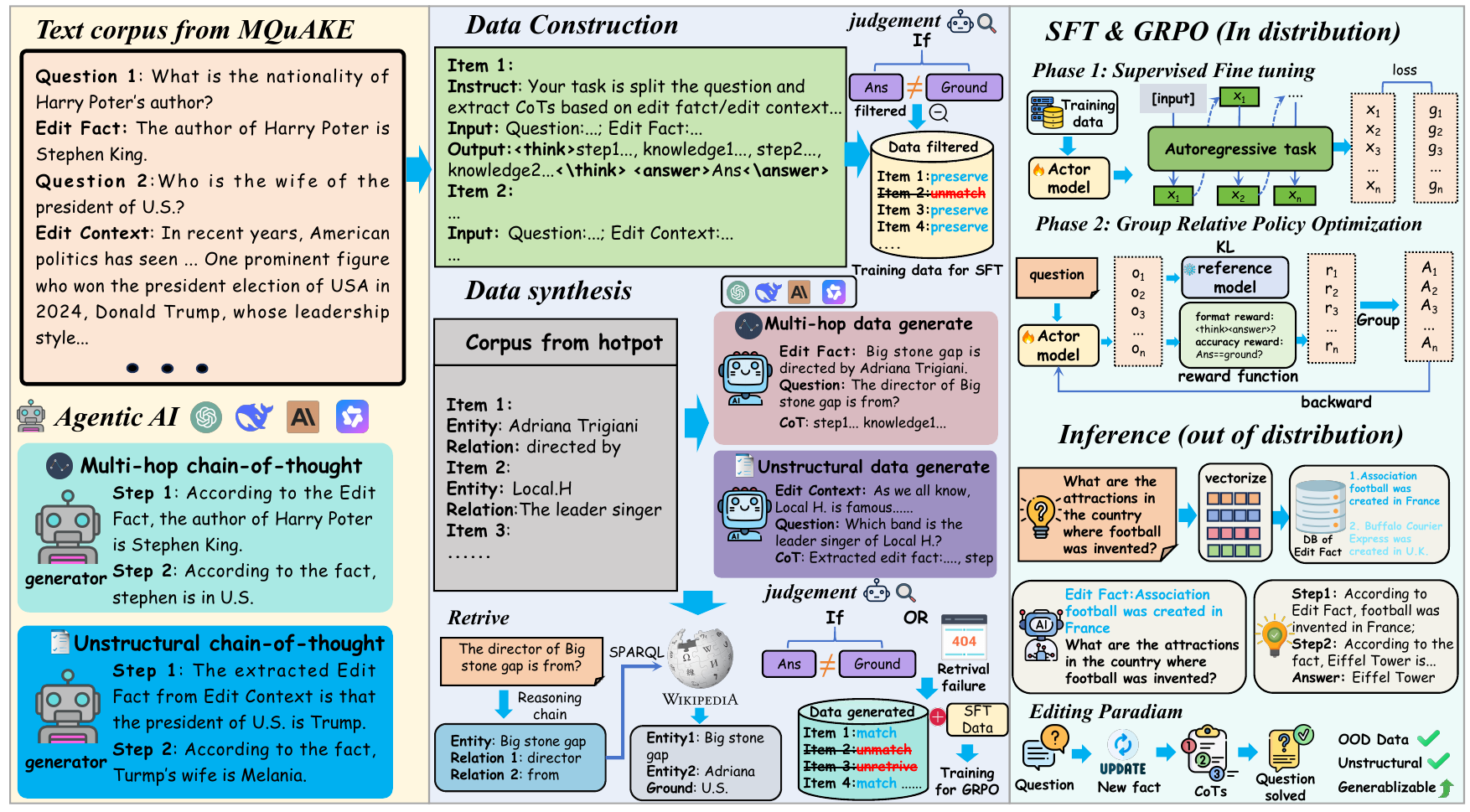}
  \caption{The framework of our proposed \llmname{CoT2Edit}. (I) Construct editing instructions by prompting LLM agents to generate reasoning chains from editing corpora. (II) Generate more instruct data by LLM agents using entity relations from HotpotQA. (III) Train the model via SFT on Phase (I) data to learn targeted response patterns. (IV) Optimize the model using GRPO to enhance generalization on the combined (I)(II) data. (V) Deploy RAG to retrieve relevant edited facts during inference, ensuring accurate question answering.
}\label{fig:2}
\end{figure*}

\section{Methods}
In this section, we propose our \llmname{CoT2Edit}, the framework of \llmname{CoT2Edit} is shown in Figure \ref{fig:2}.

\subsection{Training dataset construction}
\subsubsection{CoT Generation}
To generate reasoning chains related to editing, \myding{192} for structured data, we take the first 2,000 entries from MQuAKE \cite{zhong2023mquake} as edit instruction data; \myding{193} for unstructured data, we use all of the data in MQuAKE-uns ($\sim$ 1000 entries) to build unstructured reasoning chains, which are constructed based on the edit context and corresponding questions; \myding{194} finally, the accuracy of the generated data is checked for correctness through a result-oriented approach. 


\textit{\textbf{(I) Multi-hop Question Reasoning Chain:}} First, we construct the reasoning chain required for inference based on the edit fact $\mathcal{E}$ and question $\mathcal{Q}$ in MQuAKE. By designing a prompt template $\mathcal{T}$, we prompt a large model agent (such as \llmname{ChatGPT} or \llmname{DeepSeek}) to generate the reasoning chain. This process can be expressed as the following formula:
\begin{equation}
\label{eq.3}
Agent(\mathcal{Q}, \mathcal{E}, \mathcal{T}) \rightarrow \textbf{CoT},\mathcal{A}.
\end{equation}

By an appropriate prompt template, the solution steps can be decomposed into smaller sub-steps based on the \textbf{edit facts} and \textbf{multi-hop questions}. 

\begin{tcolorbox}[colback=blue!5!white,colframe=blue!75!black,title=Prompt template for multihop reasoning]
  \small
  Your task is to \textit{break down} the question into \textbf{steps} and extract the \textit{\textbf{chain of thought}} based on the \textbf{editing facts} into \llmname{<think></think>} tags, and finally get the corresponding answer and put it in \llmname{<answer></answer>}. You must \textbf{strictly follow} the factual information corresponding to the \textbf{Edit Facts}, a few examples of which are provided below: \{examples\}.
\end{tcolorbox}

By employing a \textbf{few-shot prompting} approach, we guide the agent to generate reasoning chains in a specific format, thereby achieving the extraction of edit facts and their associated multi-hop question-based reasoning chains.

\textit{\textbf{(II) Unstructured based reasoning chain}}: To develop a more universal knowledge editor, we further construct unstructured reasoning chains based on unstructured knowledge, enabling large language models to effectively extract relevant edit facts in response to given questions. Specifically, given an edit context $\mathcal{C}$ and a corresponding question $\mathcal{Q}$, our objective is to extract the relevant edit fact $\mathcal{E}$ from $\mathcal{C}$, then use $\mathcal{E}$ to derive the answer to $\mathcal{Q}$.
\begin{equation}
\label{eq.4}
Agent(\mathcal{Q}, \mathcal{C}, \mathcal{T}) \rightarrow \mathcal{E}, \textbf{CoT}, \mathcal{A}.
\end{equation}

\begin{tcolorbox}[colback=blue!5!white,colframe=blue!75!black,title=Prompt template for unstructured reasoning]
  \small
  Your task is to firstly \textbf{extract} the \textit{edit fact} from the \textbf{Edit Context} and secondly answer the corresponding question based on the \textit{extracted \textbf{edit fact}}, in this process you need to decompose the question and extract the \textit{chain of thought} based on the \textbf{edit fact}, put the extraction of the edit fact and the steps to decompose the question in \llmname{<think></think>} tags and put the answer in the \llmname{<answer></answer>} tags, a few examples of which are provided blow:\{examples\}.
\end{tcolorbox}

Similarly, we design appropriate prompts to allow the LLM agent to effectively extract relevant edit facts $\mathcal{E}$ and generate corresponding Chain-of-Thoughts reasoning based on the given question $\mathcal{Q}$. The prompt template $\mathcal{T}$.

Through the \textbf{few-shot prompts}, we can effectively extract the most relevant factual knowledge for editing and the CoTs required to answer the question.

\textit{\textbf{(III) Data validation}}: However, the information extracted by the LLM agent may not always be accurate. To verify the correctness of the reasoning chain, we compare the inferred answer \{$a$\} with the ground-truth answer \{$g$\} from the dataset. The consistency between these answers confirms the validity of the reasoning chain for inference tasks.

\textit{\textbf{(IV) Implementation}}: In this framework, $\mathcal{Q}$  and $\mathcal{E}$ ($\mathcal{C}$) serve as model \textbf{inputs}, while $\textbf{CoT}$ and $\mathcal{A}$ constitute the \textbf{outputs}, forming the structured instruction data for SFT. The motivation for our use of multi-hop reasoning and unstructured data is \textbf{theoretically supported} in Appendix A.3.

\subsubsection{Data Augmentation}
Since the reinforcement learning phase typically requires large-scale data for effective optimization, we employ a data augmentation approach to prompt the agent to synthesize additional instruction data.

\textbf{\myding{192} Entity-Relation Extraction:} HotpotQA \cite{hotpotqa} contains a substantial collection of fact-based entities and relations. We first extract these entity-relations from HotpotQA using heuristic methods, which serve as the foundation for constructing our augmented data. This process can be formalized as follows.
\begin{equation}
\label{eq.5}
\textbf{ent}, \textbf{rel}\leftarrow Extract(\textbf{corpus})
\end{equation}

\textbf{\myding{193} Instruct data generation:} Building upon the extracted entity-relations, we employ carefully designed prompt templates to guide the LLM agent in constructing instruction data. In this process, the agent generates corresponding questions, edit facts, and solution steps based on the identified entities and relations.
\begin{equation}
Agent(\mathcal{T}, \textbf{ent}, \textbf{rel}) \rightarrow \{\mathcal{Q},\mathcal{E},\textbf{COT},\mathcal{A}\}
\label{eq.6}
\end{equation}

The prompt $\mathcal{T}$ is described as below.

This approach utilizes key entities and relations to constrain the generated question scope, thereby preventing ambiguous or uncertain outputs while significantly improving generation accuracy.

For creating unstructured content editing data, we place the prompt template in Appendix A.1.

\textbf{\myding{194} Data validation:} Despite using carefully designed prompts, LLM agents can still generate erroneous data. To address this, we implement a three-stage filtering mechanism: (1) Entity-relation extraction from generated questions, (2) \llmname{SPARQL} based answer retrieval from Wikipedia using these extracted entities and relations as chains, (3) Automated filtering of outputs where: the generated answer mismatches the retrieved Wikipedia answer, or no valid Wikipedia reference exists. After filtering out, the scale of generated data is about 10k.

\begin{tcolorbox}[colback=blue!5!white,colframe=blue!75!black,title= Prompt template for data generation]
  \small
  You are a helpful assistant generating diverse and high-quality instruction-following data.
  1. The generated Question must require multiple steps of reasoning to solve based on the edit fact. That is to say, there are \textbf{MUST} at least two steps (Step 1, Step 2) can the question be solved.
  2. The generated Question must have the clear answer, Can't be \textbf{Unknown} or \textbf{Confused}.
  3. Can't have any extraneous strings other than json.
  4. Be sure to generate content include entity \{entities[i]\} and \{relation[i]\}.
  5. The keys in json format must \textbf{ONLY} be Input, Output.
  6. Output in the data must follow the format: The reasoning process is placed in \llmname{<think></think>} tag, the answer is placed in \llmname{<answer></answer>} tag. No other characters are allowed. A few examples:\{examples\}.
\end{tcolorbox}

\subsection{Train the model}
We utilize the data generated previously for SFT and GRPO to teach the model learning the editing paradigm. 

\textit{\textbf{Phase 1, Supervised Fine-Tuning}}. In this stage, we employ the previously constructed CoT instruction data to train the LLMs in step-by-step problem-solving using provided edit facts or edit context. This optimization process can be formally characterized as an auto-regressive task.

\begin{equation}
\mathcal{L}_{SFT}(\theta) = -\sum_{i=1}^N \sum_{t=1}^T log \mathbb{P}_\theta(y_i^t | x_i, y_i^{<t}),
\label{eq.7}
\end{equation}

where $\theta$ denotes model parameters, $D = {(x_i, y_i)}_i^N$ is the instruct dataset, $x_i$ denotes the input sequence, $y_i \in (y_i^1,y_i^2,...,y_i^T)$ denotes the target CoT sequence, and $\mathbb{P}_\theta$ denotes the conditional probability of outputting a target token given a known sequence.

Through the SFT phase, the model effectively learns a multi-step reasoning paradigm based on edit facts, thereby enabling the large language model to solve editing-related problems using CoTs reasoning.

\textit{\textbf{Phase 2, Group Relative Policy Optimization}}. In this phase, we employ GRPO to further guide behavior of LLMs by jointly leveraging both the SFT data and the synthetic data. We construct a composite reward function to guide the optimization of GRPO.
\begin{equation}
\mathcal{R} = \mathcal{R}_{acc}+\mathcal{R}_{format},
\label{eq.8}
\end{equation}
where $\mathcal{R}_{acc}$ is accuracy reward, $\mathcal{R}_{format}$ is the format reward, which is calculated by combining the inclusion of \llmname{<think><answer>} tags and keywords such as “step” or “knowledge” in the reply messages. It can effectively guide the model to generate correct reasoning paths and answers, preventing hallucinations. 

To further enhance training efficiency, we implement a self evolution strategy. After each training round, we collect high-reward samples generated during rollout and incorporate them into the next round's training dataset: 

\begin{equation} \mathcal{D}_{t+1} = \mathcal{D}_{t} \cup \{s \mid s \in \text{Rollout}_t, \mathcal{R}(s) > \theta\}, \label{eq.9} 
\end{equation} 

where $\mathcal{D}_t$ is the training dataset at round $t$, and $\theta$ is a reward threshold for selecting high-quality samples. This iterative process enables the model to rapidly learn the editing paradigm and achieve better convergence.

\begin{table*}[]
\centering
\caption{Editing performance of \llmname{CoT2Edit} against baselines on five benchmarks (editing full datasets).}
\centering
\fontsize{7}{8.25}\selectfont
\setlength{\tabcolsep}{4pt} 
\scriptsize
\begin{tabular}{c|cccccccccccc}
\hline
\rowcolor[HTML]{ECF4FF} 
\textbf{Model} & \textbf{Dataset} & \textbf{Metric} & \textbf{RoME} & \textbf{MEMIT} & \textbf{IKE} & \textbf{SKEME} & \textbf{EditCoT} & \textbf{LTE} & \textbf{PMET} & \textbf{GLAME} & \textbf{AlphaEdit} & \textbf{CoT2Edit} \\ \hline
 & \cellcolor[HTML]{EFEFEF} & \cellcolor[HTML]{EFEFEF}Edit succ. & \cellcolor[HTML]{EFEFEF}2.01 & \cellcolor[HTML]{EFEFEF}34.62 & \cellcolor[HTML]{EFEFEF}80.88 & \cellcolor[HTML]{EFEFEF}82.4 & \cellcolor[HTML]{EFEFEF}86.13 & \cellcolor[HTML]{EFEFEF}90.1 & \cellcolor[HTML]{EFEFEF}{\ul 92.63} & \cellcolor[HTML]{EFEFEF}89.48 & \cellcolor[HTML]{EFEFEF}88.78 & \cellcolor[HTML]{EFEFEF}\textbf{93.17} \\
 & \cellcolor[HTML]{EFEFEF} & \cellcolor[HTML]{EFEFEF}Para. & \cellcolor[HTML]{EFEFEF}1.8 & \cellcolor[HTML]{EFEFEF}31.28 & \cellcolor[HTML]{EFEFEF}68.13 & \cellcolor[HTML]{EFEFEF}70.27 & \cellcolor[HTML]{EFEFEF}{\ul 83.55} & \cellcolor[HTML]{EFEFEF}79.19 & \cellcolor[HTML]{EFEFEF}81.13 & \cellcolor[HTML]{EFEFEF}78.4 & \cellcolor[HTML]{EFEFEF}79.35 & \cellcolor[HTML]{EFEFEF}\textbf{92.14} \\
 & \multirow{-3}{*}{\cellcolor[HTML]{EFEFEF}ZsRE} & \cellcolor[HTML]{EFEFEF}Loc. & \cellcolor[HTML]{EFEFEF}0.69 & \cellcolor[HTML]{EFEFEF}18.49 & \cellcolor[HTML]{EFEFEF}100 & \cellcolor[HTML]{EFEFEF}100 & \cellcolor[HTML]{EFEFEF}\textbf{100} & \cellcolor[HTML]{EFEFEF}61.34 & \cellcolor[HTML]{EFEFEF}27.81 & \cellcolor[HTML]{EFEFEF}25.43 & \cellcolor[HTML]{EFEFEF}30.7 & \cellcolor[HTML]{EFEFEF}{\ul 73.13} \\
 &  & Edit succ. & 80.28 & 81.19 & 86.63 & 88.05 & 90.58 & 93.46 & 96.37 & 93.28 & {\ul 97.11} & \textbf{99.95} \\
 & \multirow{-2}{*}{MQuAKE} & Nei. & 60.38 & 64.03 & 65.38 & 67.12 & {\ul 78.71} & 74.45 & 70.28 & 77.82 & 73.63 & \textbf{92.04} \\
 & \cellcolor[HTML]{EFEFEF} & \cellcolor[HTML]{EFEFEF}Edit succ. & \cellcolor[HTML]{EFEFEF}0 & \cellcolor[HTML]{EFEFEF}0 & \cellcolor[HTML]{EFEFEF}74.97 & \cellcolor[HTML]{EFEFEF}77.35 & \cellcolor[HTML]{EFEFEF}84.64 & \cellcolor[HTML]{EFEFEF}87.47 & \cellcolor[HTML]{EFEFEF}0 & \cellcolor[HTML]{EFEFEF}0 & \cellcolor[HTML]{EFEFEF}0 & \cellcolor[HTML]{EFEFEF}\textbf{93.64} \\
 & \cellcolor[HTML]{EFEFEF} & \cellcolor[HTML]{EFEFEF}Para. & \cellcolor[HTML]{EFEFEF}0 & \cellcolor[HTML]{EFEFEF}0 & \cellcolor[HTML]{EFEFEF}65.71 & \cellcolor[HTML]{EFEFEF}70.13 & \cellcolor[HTML]{EFEFEF}{\ul 80.8} & \cellcolor[HTML]{EFEFEF}75.05 & \cellcolor[HTML]{EFEFEF}0 & \cellcolor[HTML]{EFEFEF}0 & \cellcolor[HTML]{EFEFEF}0 & \cellcolor[HTML]{EFEFEF}\textbf{79.79} \\
 & \multirow{-3}{*}{\cellcolor[HTML]{EFEFEF}Counterfact-uns} & \cellcolor[HTML]{EFEFEF}Nei. & \cellcolor[HTML]{EFEFEF}0 & \cellcolor[HTML]{EFEFEF}0 & \cellcolor[HTML]{EFEFEF}58.07 & \cellcolor[HTML]{EFEFEF}60.68 & \cellcolor[HTML]{EFEFEF}72.15 & \cellcolor[HTML]{EFEFEF}67.18 & \cellcolor[HTML]{EFEFEF}0 & \cellcolor[HTML]{EFEFEF}0 & \cellcolor[HTML]{EFEFEF}0 & \cellcolor[HTML]{EFEFEF}\textbf{93.05} \\
 &  & Edit succ. & 64.38 & 65.34 & 90.36 & 85.83 & 90.16 & 91.42 & 91.37 & 89.68 & {\ul 92.35} & \textbf{93.57} \\
 &  & Para. & 61.42 & 64.65 & 74.33 & 75.11 & {\ul 87.22} & 81.53 & 80.63 & 77.43 & 81.78 & \textbf{89.54} \\
 & \multirow{-3}{*}{Counterfact} & Nei. & 49.44 & 51.56 & 52.17 & 58.36 & {\ul 80.18} & 71.39 & 65.74 & 70.54 & 67.88 & \textbf{93.58} \\
\multirow{-12}{*}{\textbf{LLaMA3-8B}} & \cellcolor[HTML]{EFEFEF}WikiUpdate & \cellcolor[HTML]{EFEFEF}Edit succ. & \cellcolor[HTML]{EFEFEF}0 & \cellcolor[HTML]{EFEFEF}0 & \cellcolor[HTML]{EFEFEF}63.73 & \cellcolor[HTML]{EFEFEF}70.98 & \cellcolor[HTML]{EFEFEF}74.44 & \cellcolor[HTML]{EFEFEF}80.21 & \cellcolor[HTML]{EFEFEF}0 & \cellcolor[HTML]{EFEFEF}0 & \cellcolor[HTML]{EFEFEF}0 & \cellcolor[HTML]{EFEFEF}\textbf{85.89} \\ \hline
 &  & Edit succ. & 67.13 & 90.41 & 80.49 & 83.68 & 94.83 & {\ul 96.78} & 87.15 & 91.18 & 94.13 & \textbf{98.11} \\
 &  & Para. & 60.65 & 75.73 & 65.18 & 66.92 & {\ul 90.9} & 87.51 & 82.3 & 81.51 & 85.24 & \textbf{96.2} \\
 & \multirow{-3}{*}{ZsRE} & Loc. & 23.47 & 28.31 & 100 & 100 & \textbf{100} & 70.67 & 30.36 & 25.49 & 34.78 & {\ul 83.24} \\
 & \cellcolor[HTML]{EFEFEF} & \cellcolor[HTML]{EFEFEF}Edit succ. & \cellcolor[HTML]{EFEFEF}80.13 & \cellcolor[HTML]{EFEFEF}87.68 & \cellcolor[HTML]{EFEFEF}87.44 & \cellcolor[HTML]{EFEFEF}90.98 & \cellcolor[HTML]{EFEFEF}96.73 & \cellcolor[HTML]{EFEFEF}94.51 & \cellcolor[HTML]{EFEFEF}93.16 & \cellcolor[HTML]{EFEFEF}96.68 & \cellcolor[HTML]{EFEFEF}{\ul 97.73} & \cellcolor[HTML]{EFEFEF}\textbf{99.48} \\
 & \multirow{-2}{*}{\cellcolor[HTML]{EFEFEF}MQuAKE} & \cellcolor[HTML]{EFEFEF}Nei. & \cellcolor[HTML]{EFEFEF}65.21 & \cellcolor[HTML]{EFEFEF}70.34 & \cellcolor[HTML]{EFEFEF}64.29 & \cellcolor[HTML]{EFEFEF}67.43 & \cellcolor[HTML]{EFEFEF}{\ul 81.11} & \cellcolor[HTML]{EFEFEF}79.18 & \cellcolor[HTML]{EFEFEF}77.48 & \cellcolor[HTML]{EFEFEF}82.31 & \cellcolor[HTML]{EFEFEF}71.69 & \cellcolor[HTML]{EFEFEF}\textbf{92.69} \\
 &  & Edit succ. & 0 & 0 & 75.3 & 78.51 & 80.24 & {\ul 88.26} & 0 & 0 & 0 & \textbf{94.56} \\
 &  & Para. & 0 & 0 & 61.24 & 65.1 & 75.15 & {\ul 77.71} & 0 & 0 & 0 & \textbf{81.43} \\
 & \multirow{-3}{*}{Counterfact-uns} & Nei. & 0 & 0 & 61.78 & 58 & {\ul 71.11} & 65.14 & 0 & 0 & 0 & \textbf{90.61} \\
 & \cellcolor[HTML]{EFEFEF} & \cellcolor[HTML]{EFEFEF}Edit succ. & \cellcolor[HTML]{EFEFEF}67.71 & \cellcolor[HTML]{EFEFEF}87.4 & \cellcolor[HTML]{EFEFEF}82.65 & \cellcolor[HTML]{EFEFEF}84.33 & \cellcolor[HTML]{EFEFEF}90.15 & \cellcolor[HTML]{EFEFEF}{\ul 93.09} & \cellcolor[HTML]{EFEFEF}90.9 & \cellcolor[HTML]{EFEFEF}87.83 & \cellcolor[HTML]{EFEFEF}92.51 & \cellcolor[HTML]{EFEFEF}\textbf{99.13} \\
 & \cellcolor[HTML]{EFEFEF} & \cellcolor[HTML]{EFEFEF}Para. & \cellcolor[HTML]{EFEFEF}63.95 & \cellcolor[HTML]{EFEFEF}76.59 & \cellcolor[HTML]{EFEFEF}72.33 & \cellcolor[HTML]{EFEFEF}76.15 & \cellcolor[HTML]{EFEFEF}{\ul 84.48} & \cellcolor[HTML]{EFEFEF}76.17 & \cellcolor[HTML]{EFEFEF}80.23 & \cellcolor[HTML]{EFEFEF}77.78 & \cellcolor[HTML]{EFEFEF}81.63 & \cellcolor[HTML]{EFEFEF}\textbf{89.54} \\
 & \multirow{-3}{*}{\cellcolor[HTML]{EFEFEF}Counterfact} & \cellcolor[HTML]{EFEFEF}Nei. & \cellcolor[HTML]{EFEFEF}52.1 & \cellcolor[HTML]{EFEFEF}61.36 & \cellcolor[HTML]{EFEFEF}60.68 & \cellcolor[HTML]{EFEFEF}61.04 & \cellcolor[HTML]{EFEFEF}{\ul 78.55} & \cellcolor[HTML]{EFEFEF}69.66 & \cellcolor[HTML]{EFEFEF}64.35 & \cellcolor[HTML]{EFEFEF}71.82 & \cellcolor[HTML]{EFEFEF}65.77 & \cellcolor[HTML]{EFEFEF}\textbf{94.55} \\
\multirow{-12}{*}{\textbf{Falcon3-10B}} & WikiUpdate & Edit succ. & 0 & 0 & 65.36 & 71.84 & {\ul 77.18} & 74.33 & 0 & 0 & 0 & \textbf{82.29} \\ \hline
 & \cellcolor[HTML]{EFEFEF} & \cellcolor[HTML]{EFEFEF}Edit succ. & \cellcolor[HTML]{EFEFEF}59.95 & \cellcolor[HTML]{EFEFEF}90.71 & \cellcolor[HTML]{EFEFEF}86.63 & \cellcolor[HTML]{EFEFEF}85.37 & \cellcolor[HTML]{EFEFEF}91.41 & \cellcolor[HTML]{EFEFEF}90.9 & \cellcolor[HTML]{EFEFEF}94.82 & \cellcolor[HTML]{EFEFEF}92.13 & \cellcolor[HTML]{EFEFEF}95.54 & \cellcolor[HTML]{EFEFEF}\textbf{98.07} \\
 & \cellcolor[HTML]{EFEFEF} & \cellcolor[HTML]{EFEFEF}Para. & \cellcolor[HTML]{EFEFEF}54.73 & \cellcolor[HTML]{EFEFEF}78.15 & \cellcolor[HTML]{EFEFEF}67.78 & \cellcolor[HTML]{EFEFEF}63.43 & \cellcolor[HTML]{EFEFEF}85.51 & \cellcolor[HTML]{EFEFEF}79.97 & \cellcolor[HTML]{EFEFEF}87.91 & \cellcolor[HTML]{EFEFEF}84.95 & \cellcolor[HTML]{EFEFEF}90.45 & \cellcolor[HTML]{EFEFEF}\textbf{96.64} \\
 & \multirow{-3}{*}{\cellcolor[HTML]{EFEFEF}ZsRE} & \cellcolor[HTML]{EFEFEF}Loc. & \cellcolor[HTML]{EFEFEF}20.23 & \cellcolor[HTML]{EFEFEF}23.33 & \cellcolor[HTML]{EFEFEF}100 & \cellcolor[HTML]{EFEFEF}100 & \cellcolor[HTML]{EFEFEF}\textbf{100} & \cellcolor[HTML]{EFEFEF}73.63 & \cellcolor[HTML]{EFEFEF}26.61 & \cellcolor[HTML]{EFEFEF}24.04 & \cellcolor[HTML]{EFEFEF}36.37 & \cellcolor[HTML]{EFEFEF}{\ul 85.59} \\
 &  & Edit succ. & 80.74 & 91.65 & 84.93 & 87.28 & 96.37 & 97.73 & 94.13 & 95.56 & 97.71 & \textbf{99.22} \\
 & \multirow{-2}{*}{MQuAKE} & Nei. & 70.15 & 73.48 & 64.29 & 66.13 & {\ul 83.82} & 77.65 & 79.78 & 84.31 & 76.67 & \textbf{92.24} \\
 & \cellcolor[HTML]{EFEFEF} & \cellcolor[HTML]{EFEFEF}Edit succ. & \cellcolor[HTML]{EFEFEF}0 & \cellcolor[HTML]{EFEFEF}0 & \cellcolor[HTML]{EFEFEF}80.38 & \cellcolor[HTML]{EFEFEF}82.85 & \cellcolor[HTML]{EFEFEF}87.76 & \cellcolor[HTML]{EFEFEF}91.48 & \cellcolor[HTML]{EFEFEF}0 & \cellcolor[HTML]{EFEFEF}0 & \cellcolor[HTML]{EFEFEF}0 & \cellcolor[HTML]{EFEFEF}\textbf{94.46} \\
 & \cellcolor[HTML]{EFEFEF} & \cellcolor[HTML]{EFEFEF}Para. & \cellcolor[HTML]{EFEFEF}0 & \cellcolor[HTML]{EFEFEF}0 & \cellcolor[HTML]{EFEFEF}52.63 & \cellcolor[HTML]{EFEFEF}56.18 & \cellcolor[HTML]{EFEFEF}80.53 & \cellcolor[HTML]{EFEFEF}74.75 & \cellcolor[HTML]{EFEFEF}0 & \cellcolor[HTML]{EFEFEF}0 & \cellcolor[HTML]{EFEFEF}0 & \cellcolor[HTML]{EFEFEF}\textbf{81.02} \\
 & \multirow{-3}{*}{\cellcolor[HTML]{EFEFEF}Counterfact-uns} & \cellcolor[HTML]{EFEFEF}Nei. & \cellcolor[HTML]{EFEFEF}0 & \cellcolor[HTML]{EFEFEF}0 & \cellcolor[HTML]{EFEFEF}53.37 & \cellcolor[HTML]{EFEFEF}55.7 & \cellcolor[HTML]{EFEFEF}{\ul 76.06} & \cellcolor[HTML]{EFEFEF}63.15 & \cellcolor[HTML]{EFEFEF}0 & \cellcolor[HTML]{EFEFEF}0 & \cellcolor[HTML]{EFEFEF}0 & \cellcolor[HTML]{EFEFEF}\textbf{84.51} \\
 &  & Edit succ. & 63.07 & 92.13 & 84.43 & 82.58 & {\ul 96.73} & 95.31 & 96.63 & 92.44 & 95.57 & \textbf{99.04} \\
 &  & Para. & 55.33 & 80.35 & 67.14 & 69.37 & {\ul 90.9} & 87.17 & 86.37 & 81.47 & 87.25 & \textbf{98.94} \\
 & \multirow{-3}{*}{Counterfact} & Nei. & 51.23 & 66.47 & 60.35 & 63.08 & {\ul 81.16} & 78.56 & 74.31 & 78.73 & 68.54 & \textbf{92.07} \\
\multirow{-12}{*}{\textbf{Qwen3-14B}} & \cellcolor[HTML]{EFEFEF}WikiUpdate & \cellcolor[HTML]{EFEFEF}Edit succ. & \cellcolor[HTML]{EFEFEF}0 & \cellcolor[HTML]{EFEFEF}0 & \cellcolor[HTML]{EFEFEF}72.03 & \cellcolor[HTML]{EFEFEF}69.25 & \cellcolor[HTML]{EFEFEF}{\ul 77.68} & \cellcolor[HTML]{EFEFEF}75.03 & \cellcolor[HTML]{EFEFEF}0 & \cellcolor[HTML]{EFEFEF}0 & \cellcolor[HTML]{EFEFEF}0 & \cellcolor[HTML]{EFEFEF}\textbf{82.76} \\ \hline
\end{tabular}
\label{tab:1}
\end{table*}

\subsection{Inference}
To achieve more practical on-the-fly editing, we incorporate RAG to retrieve the relevant edit fact with the question. We utilize an off-the-shelf retrieval model \llmname{contriver-msmarco} invented by Facebook to embed all the edit facts and create a vector memory to store the representations. When given a query, we also get the representation of the query by the retriever and search the most similar edit fact from the vector memory. Then, the query and retrieved edit fact are fed into the LLM to obtain the reasoning chains and answers.

\section{Experiments}
In this section, we perform a wide range of experiments to answer the following questions.

\textbf{RQ1}: How does \llmname{CoT2Edit} perform on sequential editing tasks compared to baselines in five structural and unstructured editing benchmarks?

\textbf{RQ2}: How does the out-of-distribution capability of \llmname{CoT2Edit} in unseen editing task?

\textbf{RQ3}: How does the modular design of \llmname{CoT2Edit} enhance the model's capabilities?

\textbf{RQ4}: How does the performance of \llmname{CoT2Edit} in practical editing scenarios? How can we evaluate its actual effectiveness?

\subsection{Experimental setup}
\textbf{LLMs and Baseline methods}.  Our experiments are carried out on three base LLMs: \llmname{Llama3-8B} \cite{grattafiori2024llama}, \llmname{Falcon3-10B} \cite{almazrouei2023falcon}, and \llmname{Qwen3-14B} \cite{yang2025qwen3}. We compare our method against \llmname{Fine-tuning} \cite{hu2021lora}, \llmname{RoME} \cite{meng2022locating}, \llmname{MEMIT} \cite{meng2022mass}, \llmname{IKE} \cite{zheng2023can}, \llmname{SKEME } \cite{chen-etal-2024-robust} \llmname{PMET} \cite{zhang2024knowledge}, and \llmname{GLAME} \cite{zhang2024knowledge} and  \llmname{AlphaEdit} \cite{fang2024alphaedit}. More details are given in Appendix B.3.

\textbf{Datasets}. We evaluate \llmname{CoT2Edit} using three structural benchmarks and two unstructured benchmarks. CounterFact \cite{meng2022mass}, ZsRE \cite{levy2017zero} and MQuAKE \cite{zhong2023mquake} are structural datasets; CounterFact-uns and WikiUpdate \cite{AKEW} are unstructured datasets. Edit success as Edit succ., Paraphrase as Para., Locality as Loc., and Neighborhood success as Nei are adopted as metrics \cite{meng2022mass}. More details are provided in Appendix B.1, B.2, and B.4.

\subsection{Editing performance (RQ1)}
We conduct editing experiments on three based LLMs using five datasets. Based on Table \ref{tab:1}, we can obtain the following observation.

\textbf{Obs 1:} \textbf{\llmname{CoT2Edit} achieves superior performance in nearly all metrics and base models.} From Table \ref{tab:1}, we can observe that \llmname{CoT2Edit} provides the best editing success performance in all three base models. This demonstrates that training and reasoning through CoTs not only enhances the model’s confidence in the edited facts but also ensures that the entire process remains fully interpretable.

\textbf{Obs 2:} \textbf{\llmname{CoT2Edit} has a stronger paraphrase and multi-hop inference ability against advance baselines.} In the paraphrase metric, it's about $\uparrow\sim 10\%$ compared to \llmname{EditCoT}. In the neighbor success metric, it's most $\sim \uparrow 15\%$ compared to the strongest \llmname{EditCoT} and \llmname{GLAME} in MQuAKE. Unlike prior methods that often rely on rote memorization, \llmname{CoT2Edit} effectively transfers edited knowledge to semantically diverse and contextually related queries, demonstrating stronger robustness under rephrased and multi-hop conditions.

\textbf{Obs 3:} \textbf{\llmname{CoT2Edit} possesses strong capabilities to edit unstructured knowledge.} Parameter-editing  methods are inherently limited to modifying knowledge expressed in the form of structured triples. As a result, they failed to edit free-form textual knowledge. \llmname{CoT2Edit} achieves the best performance in unstructured knowledge editing in all three models, maintaining an accuracy greater than 80\% ($\uparrow 5-15\%$ in three base models) even in cross-lingual knowledge editing scenarios such as WikiUpdate.

\subsection{Out-of-distribution Generalization (RQ2)}
To evaluate \llmname{CoT2Edit} performance in OOD scenarios, we conduct rigorous experiments on ConvSent \cite{mitchell2022memory}, a sentiment editing task featuring diverse data distributions, alongside established benchmarks.

\begin{table}[]
\centering
\caption{OOD generalization of \llmname{CoT2Edit} on ConvSent bench using \llmname{Llama3-8B} and \llmname{Falcon-3-10B}.}
\scriptsize
\setlength{\tabcolsep}{2.2pt}
\begin{tabular}{c|cccccc}
\hline
\rowcolor[HTML]{FFFFFF} 
{\color[HTML]{2C2C36} \textbf{Models}} & {\color[HTML]{2C2C36} \textbf{Methods}} & {\color[HTML]{2C2C36} \textbf{1K Edits}} & {\color[HTML]{2C2C36} \textbf{3K Edits}} & {\color[HTML]{2C2C36} \textbf{5K Edits}} & {\color[HTML]{2C2C36} \textbf{8K Edits}} & {\color[HTML]{2C2C36} \textbf{10K Edits}} \\ \hline
\rowcolor[HTML]{FFFFFF} 
\cellcolor[HTML]{EFEFEF} & {\color[HTML]{2C2C36} MEMIT} & {\color[HTML]{2C2C36} 58.72} & {\color[HTML]{2C2C36} 54.54} & {\color[HTML]{2C2C36} 50.66} & {\color[HTML]{2C2C36} 46.9} & {\color[HTML]{2C2C36} 43.06} \\
\rowcolor[HTML]{FFFFFF} 
\cellcolor[HTML]{EFEFEF} & {\color[HTML]{2C2C36} IKE} & {\color[HTML]{2C2C36} 41.19} & {\color[HTML]{2C2C36} 38.32} & {\color[HTML]{2C2C36} 36.2} & {\color[HTML]{2C2C36} 32.37} & {\color[HTML]{2C2C36} 29.33} \\
\rowcolor[HTML]{FFFFFF} 
\cellcolor[HTML]{EFEFEF} & {\color[HTML]{2C2C36} AlphaEdit} & {\color[HTML]{2C2C36} 76.13} & {\color[HTML]{2C2C36} 70.53} & {\color[HTML]{2C2C36} 63.28} & {\color[HTML]{2C2C36} 55.38} & {\color[HTML]{2C2C36} 53.11} \\
\rowcolor[HTML]{FFFFFF} 
\cellcolor[HTML]{EFEFEF} & {\color[HTML]{2C2C36} PMET} & {\color[HTML]{2C2C36} 67.13} & {\color[HTML]{2C2C36} 65.11} & {\color[HTML]{2C2C36} 63.27} & {\color[HTML]{2C2C36} 61.15} & {\color[HTML]{2C2C36} 60.58} \\
\rowcolor[HTML]{FFFFFF} 
\cellcolor[HTML]{EFEFEF} & {\color[HTML]{2C2C36} LTE} & {\color[HTML]{2C2C36} 80.49} & {\color[HTML]{2C2C36} 78.83} & {\color[HTML]{2C2C36} 76.11} & {\color[HTML]{2C2C36} 73.37} & {\color[HTML]{2C2C36} 72.78} \\
\rowcolor[HTML]{FFFFFF} 
\cellcolor[HTML]{EFEFEF} & {\color[HTML]{2C2C36} EditCoT} & {\color[HTML]{2C2C36} 78.34} & {\color[HTML]{2C2C36} 76.79} & {\color[HTML]{2C2C36} 76.43} & {\color[HTML]{2C2C36} 75.11} & {\color[HTML]{2C2C36} 74.91} \\
\rowcolor[HTML]{CBCEFB} 
\multirow{-5}{*}{\cellcolor[HTML]{EFEFEF}Llama3} & {\color[HTML]{2C2C36} \textbf{COT2Edit}} & {\color[HTML]{2C2C36} \textbf{87.25}} & {\color[HTML]{2C2C36} \textbf{86.18}} & {\color[HTML]{2C2C36} \textbf{85.33}} & {\color[HTML]{2C2C36} \textbf{85.27}} & {\color[HTML]{2C2C36} \textbf{85.2}} \\ \hline
\rowcolor[HTML]{FFFFFF} 
\cellcolor[HTML]{EFEFEF} & {\color[HTML]{2C2C36} MEMIT} & {\color[HTML]{2C2C36} 61.48} & {\color[HTML]{2C2C36} 57.72} & {\color[HTML]{2C2C36} 53.17} & {\color[HTML]{2C2C36} 50.58} & {\color[HTML]{2C2C36} 47.36} \\
\rowcolor[HTML]{FFFFFF} 
\cellcolor[HTML]{EFEFEF} & {\color[HTML]{2C2C36} IKE} & {\color[HTML]{2C2C36} 40.75} & {\color[HTML]{2C2C36} 36.53} & {\color[HTML]{2C2C36} 33.45} & {\color[HTML]{2C2C36} 31.08} & {\color[HTML]{2C2C36} 27.99} \\
\rowcolor[HTML]{FFFFFF} 
\cellcolor[HTML]{EFEFEF} & {\color[HTML]{2C2C36} AlphaEdit} & {\color[HTML]{2C2C36} 78.07} & {\color[HTML]{2C2C36} 77.75} & {\color[HTML]{2C2C36} 74.63} & {\color[HTML]{2C2C36} 72.18} & {\color[HTML]{2C2C36} 70.7} \\
\rowcolor[HTML]{FFFFFF} 
\cellcolor[HTML]{EFEFEF} & {\color[HTML]{2C2C36} PMET} & {\color[HTML]{2C2C36} 71.18} & {\color[HTML]{2C2C36} 70.03} & {\color[HTML]{2C2C36} 69.67} & {\color[HTML]{2C2C36} 68.13} & {\color[HTML]{2C2C36} 66.55} \\
\rowcolor[HTML]{FFFFFF} 
\cellcolor[HTML]{EFEFEF} & {\color[HTML]{2C2C36} LTE} & {\color[HTML]{2C2C36} 77.37} & {\color[HTML]{2C2C36} 76.15} & {\color[HTML]{2C2C36} 75.76} & {\color[HTML]{2C2C36} 72.62} & {\color[HTML]{2C2C36} 71.03} \\
\rowcolor[HTML]{FFFFFF} 
\cellcolor[HTML]{EFEFEF} & {\color[HTML]{2C2C36} EditCoT} & {\color[HTML]{2C2C36} 79.15} & {\color[HTML]{2C2C36} 78.81} & {\color[HTML]{2C2C36} 76.12} & {\color[HTML]{2C2C36} 75.27} & {\color[HTML]{2C2C36} 74.51} \\
\rowcolor[HTML]{CBCEFB} 
\multirow{-5}{*}{\cellcolor[HTML]{EFEFEF}Falcon3} & {\color[HTML]{2C2C36} \textbf{COT2Edit}} & {\color[HTML]{2C2C36} \textbf{84.32}} & {\color[HTML]{2C2C36} \textbf{83.01}} & {\color[HTML]{2C2C36} \textbf{82.82}} & {\color[HTML]{2C2C36} \textbf{80.68}} & {\color[HTML]{2C2C36} \textbf{79.08}} \\ \hline
\end{tabular}
\label{tab:3}
\end{table}

\textbf{Obs 4:} \textbf{\llmname{CoT2Edit} has the strongest generalization of OOD.} As shown in Table \ref{tab:3}, our \llmname{CoT2Edit} outperforms existing methods at least $\uparrow 11\%$ on unseen sentiment editing tasks. In contrast, locate-then-edit approaches exhibit poor generalization to OOD cases (merely at most 76\%-78\%), with performance worsening as the number of edits increases (decrease 10-20\%). Context learning methods, struggle to capture sentiment changes using only prompt templates, further highlighting the advantage of \llmname{CoT2Edit}. The interpretable of \llmname{CoT2Edit} superior in this task is also explained by case studies in Appendix B.5. 
\subsection{Ablation study (RQ3)}

In this section, an ablation study is conducted, as shown in Table \ref{tab:2}. More datasets are shown in Appendix B.5.

\begin{table}[htbp]
\caption{Ablation studies of \llmname{CoT2Edit} on CounterFact.}
\centering
\scriptsize
\setlength{\tabcolsep}{4pt}
\begin{tabular}{c|cccc}
\hline
{\color[HTML]{2C2C36} \textbf{Model}} & \textbf{Varient} & \textbf{Edit Succ.} & \textbf{Para.} & \textbf{Nei.} \\ \hline
 & CoT2Edit-W/O SFT & 86.8 & 83.28 & 80.1 \\
 & CoT2Edit-W/O GRPO & 90.95 & 84.03 & 87.38 \\
 & CoT2Edit-W/O Train & 55.11 & 50.5 & 47.18 \\
\multirow{-4}{*}{Llama-3-8B} & \cellcolor[HTML]{DAE8FC}CoT2Edit & \cellcolor[HTML]{DAE8FC}\textbf{93.57} & \cellcolor[HTML]{DAE8FC}\textbf{89.54} & \cellcolor[HTML]{DAE8FC}\textbf{92.76} \\ \hline
 & CoT2Edit-W/O SFT & 92.71 & 85.63 & 84.37 \\
 & CoT2Edit-W/O GRPO & 96.93 & 86.34 & 87.78 \\
 & CoT2Edit-W/O Train & 57.31 & 53.68 & 50.93 \\
\multirow{-4}{*}{Falcon3-10B} & \cellcolor[HTML]{DAE8FC}CoT2Edit & \cellcolor[HTML]{DAE8FC}\textbf{99.13} & \cellcolor[HTML]{DAE8FC}\textbf{89.54} & \cellcolor[HTML]{DAE8FC}\textbf{94.55} \\ \hline
 & CoT2Edit-W/O SFT & 87.13 & 84.42 & 85.5 \\
 & CoT2Edit-W/O GRPO & 92.06 & 90.98 & 87.11 \\
 & CoT2Edit-W/O Train & 61.73 & 56.56 & 57.81 \\
\multirow{-4}{*}{Qwen3-14B} & \cellcolor[HTML]{DAE8FC}CoT2Edit & \cellcolor[HTML]{DAE8FC}\textbf{99.04} & \cellcolor[HTML]{DAE8FC}\textbf{98.94} & \cellcolor[HTML]{DAE8FC}\textbf{92.07} \\ \hline
\end{tabular}
\label{tab:2}
\end{table}
\textbf{Obs 5:} \textbf{SFT and GRPO are two indispensable steps.} \textit{Without SFT}, the editing performance drops by around 7\% in general. This shows that SFT helps the model quickly learn the basic reasoning format and task structure, even with limited data. \textit{Without GRPO}, the model also shows clear performance degradation, particularly in Para. and Nei. metrics($\downarrow 5-8\%$), highlighting that GRPO improves factual accuracy and generalization through reward optimization.

\subsection{Practicality analysis (RQ4)}
In this section, we demonstrate the practicality of \llmname{CoT2Edit} from four perspectives. 

\textbf{\myding{182} Time consumption}: We analyze the time consumption of our methods during the inference time. The results are shown in Table \ref{tab:4}.
\begin{table}[htbp]
\caption{Averaged Wall Clock Time per edit method
for 10 edits on ZsRE using \llmname{LLaMA3-8B}.}
\small
\begin{tabular}{cccc}
\hline
\textbf{Method} & \textbf{Edit Time} & \textbf{Inference Time} & \textbf{Total Time} \\ \hline
FT & 62.73 & 1.41 & 64.14 \\
IKE & 0.00 & 1.71 & \cellcolor[HTML]{9698ED}1.71 \\
RoME & 210.77 & 1.55 & 213.32 \\
MEMIT & 183.24 & 1.57 & 184.81 \\
LTE & 0.00 & 1.60 & \cellcolor[HTML]{9698ED}1.60 \\
AlphaEdit & 199.13 & 1.6 & 200.03 \\
EditCoT & 0.00     & 4.07  & 4.07 \\
CoT2Edit & 0.00 & 1.93 & \cellcolor[HTML]{9698ED}1.93 \\ \hline
\end{tabular}
\label{tab:4}
\end{table}

\textbf{Obs 6: \llmname{CoT2Edit} stands out by achieving the swiftest editing speeds coupled with superior performance.} Methods like \llmname{RoME} and \llmname{MEMIT}, which follow a locate-then-edit approach, require opening the model parameters for editing and then performing inference whenever new knowledge is introduced. This makes the editing process quite complex. After just a single train round (which takes at most 4 hours in our experiments compare 9 hours of \llmname{LTE}), \llmname{CoT2Edit} enables instantaneous editing similar to \llmname{IKE} and \llmname{LTE} by appending a knowledge editing prompt to the input prefix. Despite a marginally increased inference time, the overall time expenditure is significantly reduced, underscoring the efficiency and effectiveness of \llmname{CoT2Edit}. Due to the complex pipeline of \llmname{EditCoT}, its inference time is more than twice longer than that of \llmname{CoT2Edit}.

\begin{figure}[htbp]
  \centering
  \includegraphics[height=2.5cm]{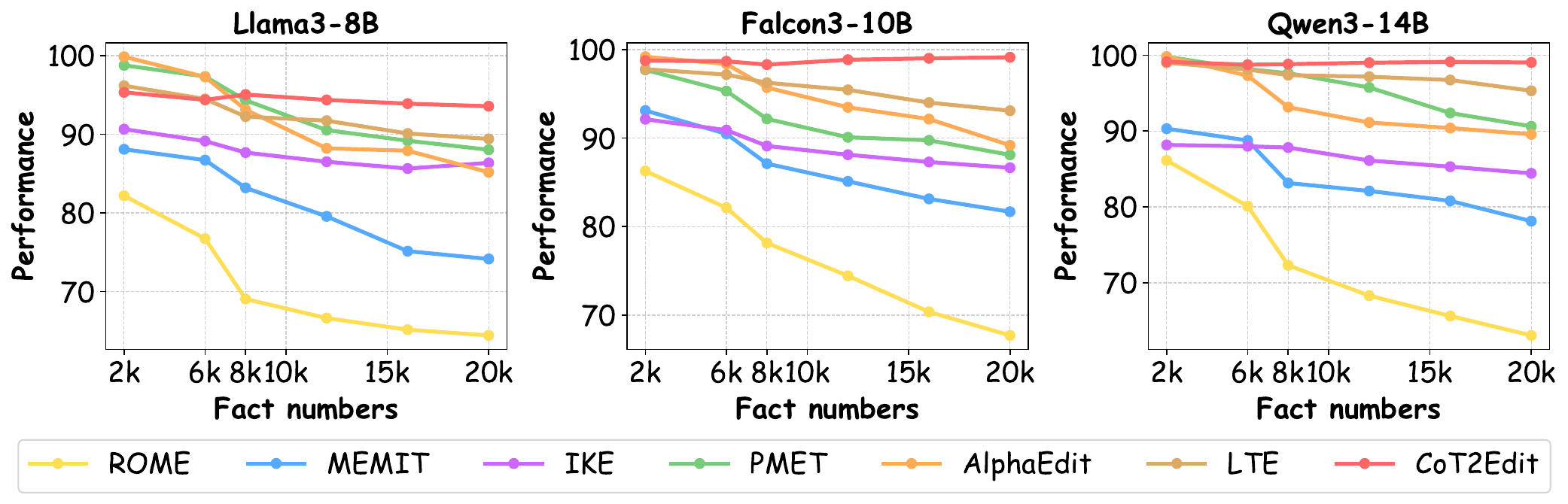}
  \caption{The impact of factual number on editing efficacy in CounterFact.}\label{fig:3}
  \vspace{-5mm}
\end{figure}

\textbf{\myding{183} Edit number}: We analyze the influence of editing number on performance through Figure \ref{fig:3}.

\textbf{Obs 7: The editing effect of \llmname{CoT2Edit} does not collapse as the number of facts increases.} Due to cumulative errors, as the number of edited facts increases, the performance of the locate-then-edit methods gradually deteriorates. However, since \llmname{CoT2Edit} employs a train-inference framework, its effectiveness does not gradually diminish as the number of edited facts increases. Since \llmname{LTE} lacks the support of an inference path, it achieves suboptimal results.

\begin{figure}[htbp]
  \centering
  \includegraphics[height=3.2cm]{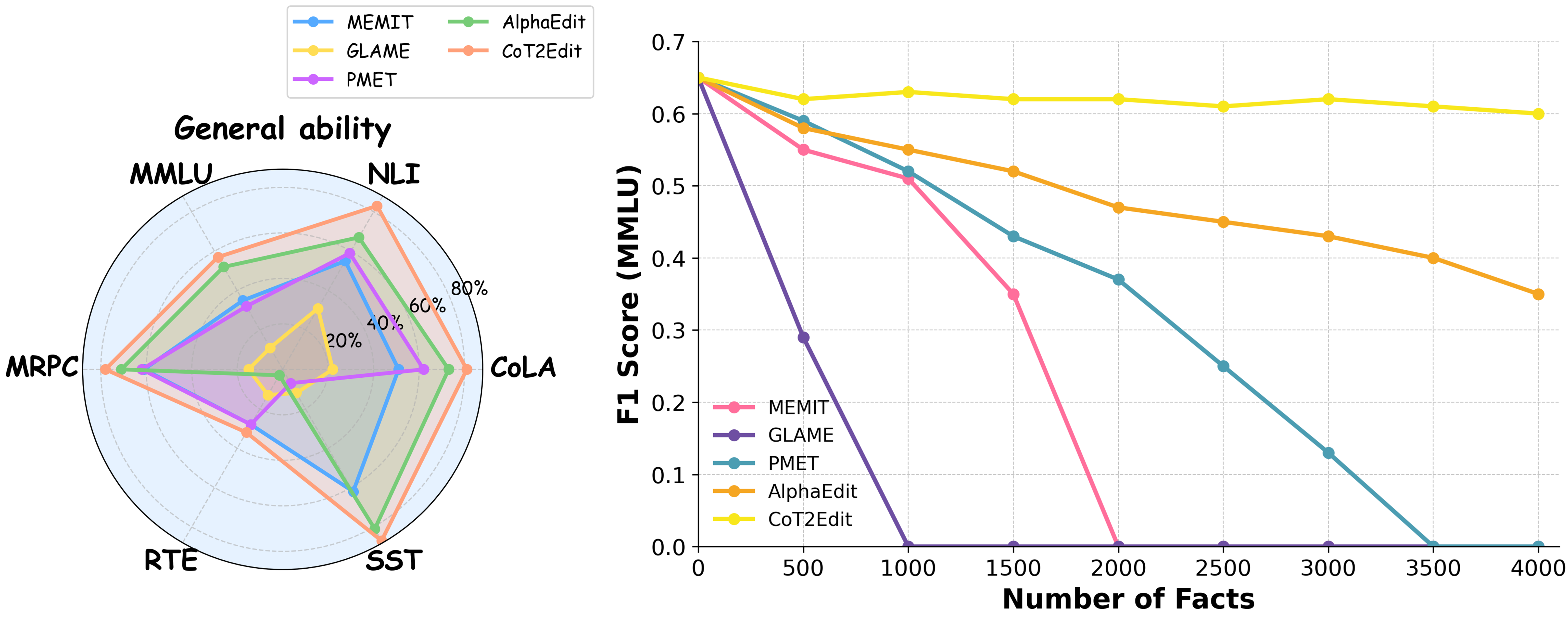}
  \caption{The evaluation of general ability of LLM in \llmname{Llama-3-8B}.}\label{fig:4}
  \vspace{-5mm}
\end{figure}

\textbf{\myding{184} General ability}: Figure \ref{fig:4} illustrates the general performance metrics of the model when editing \llmname{Llama3-8b}. 

\textbf{Obs 8: \llmname{CoT2Edit} retains the highest level of original performance from LLMs.} \llmname{CoT2Edit} achieved the best performance in the six benchmarks evaluating general performance such as SST \cite{socher2013recursive}, CoLA \cite{warstadt2019neural} etc. As shown on the right side of Figure 4, even though \llmname{AlphaEdit} uses a null space method, its normal performance deteriorates on MMLU \cite{hendrycks2020measuring} due to accumulated errors as the number of facts increases. In contrast, \llmname{CoT2Edit} maintains stable performance as the editing scale grows.


  
  
  

\section{Conclusions}

In this paper, we propose a novel editing paradigm \llmname{CoT2Edit} that enables LLMs to perform knowledge editing in real-world QA scenarios.  We demonstrate the generality of \llmname{CoT2Edit} through experiments on six benchmarks and extensively evaluate efficiency, OOD performance, and its general capabilities. In future work, we will construct more diverse instruction datasets to enhance the adaptability of \llmname{CoT2Edit} to more scenarios. 

\section{Limitations}
Our instruction dataset currently consists only of multi-hop and unstructured questions, which may not be sufficient to cover a broader range of editing scenarios. Future work should focus on constructing more diverse and comprehensive editing instruction sets to enhance the adaptability of the method.

\bibliography{custom}

\appendix

\section{Methodological Supplements}
\subsection{Prompt template for unstructured data generation}
To achieve superior performance in the unstructured editing task, we generate more instruction data about it. HotpotQA \cite{hotpotqa} is also used as corpus for data generation. We extract the entity-relations from HotpotQA using heuristic methods by equation (5), and then design a specific prompt to make LLM agent to generate instruct data. This process can be formulated as:
\begin{equation}
\{\mathcal{Q},\mathcal{C},\textbf{COT},\mathcal{A}\}\rightarrow Agent(\mathcal{T}, \textbf{ent}, \textbf{rel}),
\label{eq.1}
\end{equation}
where $\mathcal{C}$ denotes the constructed edit fact. The prompt template $\mathcal{T}$ are as follows.
\begin{tcolorbox}[colback=blue!5!white,colframe=blue!75!black,title= Agent prompt template]
  \small
  You are a helpful assistant generating diverse and high-quality instruction-following data.
  1. Use creative but plausible edit contexts and questions.
  2. The generated Question must have the clear answer, Can't be \textbf{Unknown} or \textbf{Confused}.
  3. Can't have any extraneous strings other than json.
  4. The format of the example data must be followed.
  5. Be sure to generate content about entity {entities[i]} and {relation[i]}.
  6. The keys in json format must \textbf{ONLY} be Input, Output
  7. Output in the data must follow the format: The reasoning process is placed in \llmname{<think></think>} tag, the answer is placed in \llmname{<answer></answer>} tag. No other characters are allowed. A few examples:\{examples\}.
\end{tcolorbox}
This approach utilizes key entities and relations to constrain the generated question scope, thereby preventing ambiguous or uncertain outputs while significantly improving generation accuracy. By employing few-shot prompting, we guide the LLM agent to recognize that factual information must be implicitly embedded within the editing context, and that relevant factual knowledge should be effectively extracted during the process of chain-of-thought reasoning.

\subsection{Prompt template for Convsent}
To better adapt our method to the ConvSent dataset, we designed specific instruct that enable our editing paradigm to generalize to previously unseen editing tasks.
\begin{tcolorbox}[colback=blue!5!white,colframe=blue!75!black,title=Instruct template for Convsent]
  \small
  Your task is to reason \textit{\textbf{step-by-step}} based on the \textbf{entity} and \textbf{edit sentiment}, and then give your sentiment about the entity. Put your reasoning inside \texttt{<think></think>} and your final sentiment about the given entity in \texttt{<answer></answer>}. You \textbf{Must} follow the sentiment in the sentence of \textbf{Edit Sentiment}.
\end{tcolorbox}

By slightly modifying the instructions used in our factual editing tasks, our method can rapidly generalize to unseen editing tasks and achieve stronger performance against baselines.

\subsection{Theoretical motivation about training data construction}
This section presents the motivation for employing \textbf{multi-hop reasoning} and \textbf{unstructured data} in model training, along with the theoretical underpinnings supporting this design.

\textbf{\ding{224} Multi-hop Reasoning} 

The fundamental challenge in knowledge editing extends beyond mere fact injection, it requires models to develop robust reasoning capabilities that can generalize across diverse query formulations and contexts. We argue that multi-hop reasoning training provides a principled approach to address the limitations of existing parameter-modification methods, which primarily rely on superficial key-value associations.

\ding{182} \textbf{Formal framework for multi-hop knowledge integration}

Consider a knowledge editing scenario where we aim to learn a mapping function $f_{\theta'}: \mathcal{E} \times \mathcal{Q} \rightarrow \mathcal{A}$, where $\mathcal{E}$ denotes the space of editable facts, $\mathcal{Q}$ represents possible queries, and $\mathcal{A}$ denotes the answer space. Traditional locate-then-edit methods \cite{meng2022mass} optimize this function through direct parameter modification:

\begin{equation} \small
\label{eq.7}
\textbf{W}_{out}\overset{\Delta}{=}\underset{\hat{\textbf{W}}_{out}}{\rm{argmin}}(||\hat{\textbf{W}}_{out}\textbf{K}_1-\textbf{V}_1||^2+||(\hat{\textbf{W}}_{out}\textbf{K}_0-\textbf{V}_0||^2),
\end{equation}

However, this approach treats knowledge as atomic key-value pairs, fundamentally limiting the model's ability to perform compositional reasoning. In contrast, multi-hop reasoning decomposes complex queries into a sequence of interconnected sub-problems:

\begin{equation}
\label{eq.8}
\mathcal{Q}_m=\{q|q=compose(q_1,q_2,...,q_n)\},
\end{equation}
where $compose(\cdot)$ represents the compositional operation that chains individual reasoning steps, and each $q_i$ corresponds to a specific inferential component.

\ding{183} \textbf{Theoretical advantages of multi-hop training}

\begin{dinglist}{117}

\item \textbf{Enhanced Semantic Robustness through Compositional Learning}. Multi-hop reasoning training addresses the phrasing sensitivity problem by forcing models to learn deeper semantic relationships rather than surface-level pattern matching \cite{biran-etal-2024-hopping}. This is grounded in compositional generalization theory \cite{min2019compositional}, which posits that structured learning should enable connectionist systems to generalize in more predictable and systematic ways. 

Formally, let $\phi: \mathcal{K} \rightarrow \mathcal{H}$ be a representation function that maps knowledge $\mathcal{K}$ to a hidden representation space $\mathcal{H}$. Multi-hop training ensures that:

\begin{equation}
\label{eq.9}
\phi(extract(c))\approx \phi((s,r,o))
\end{equation}
where context $c$ contains the factual triple $(s,r,o)$, thereby achieving representation invariance across different surface forms.

\item \textbf{Cognitive Load Distribution and Generalization}. The cognitive load distribution theory \cite{xu2024large} suggests that multi-hop reasoning training redistributes the model's computational burden from rote memorization to pattern-based inference. This can be mathematically expressed as:
\begin{equation}
\label{eq.10} \small
\mathcal{L}_{mem}(k, m) = \sum_i weight(k_i) \times memorize(k_i, m)
\end{equation}
\begin{equation}
\label{eq.11} \small
\mathcal{L}_{r}(k, m) = \sum_i weight(k_i) \times reason(k_i, m)
\end{equation}
where $\mathcal{L}(\cdot)$ is the cognitive load function, $\mathcal{L}_{mem}$ is the traditional knowledge pattern emphasizing memorization, $\mathcal{L}_{r}$ reflects the load under multi-hop reasoning, $k_i$ denotes a knowledge unit, $m$ is the model, and $weight(k_i)$ quantifies the relevance or frequency of $k_i$. 

Empirically, reason($k_i$, m) < memorize($k_i$, m) for a well-trained model, since reasoning over compositional structures typically requires less cognitive load to generalize across diverse inputs compared to memorizing discrete facts \cite{biran-etal-2024-hopping}. This shift in cognitive load distribution redirects the model's computational focus from rote memorization of specific facts to pattern-based reasoning, thereby enhancing its generalization capability for multiple editing tasks.

\item \textbf{Information-Theoretic Justification}. From an information-theoretic perspective, multi-hop reasoning training reduces the minimum description length (MDL) \cite{hansen2001model} of the model by learning compressed representations of reasoning patterns rather than storing individual facts. The total description length is:

\begin{equation} \small
\label{eq.12}
\mathcal{L}_{total} = \mathcal{C}(M)+\mathcal{F}(D|M)
\end{equation}

where $\mathcal{C}(M)$ represents model complexity and $\mathcal{F}(D|M)$ represents data fitting cost. Multi-hop training effectively reduces $\mathcal{C}(M)$ by learning generalizable reasoning schemas that can be applied across diverse scenarios.

\end{dinglist}

\textbf{\ding{224} Unstructured data-guided training}

Real-world knowledge rarely manifests in the rigid triple format $(s,r,o)$ assumed by existing editing methods \cite{AKEW}. Instead, factual information is embedded within free-form text, requiring models to perform knowledge extraction and reasoning simultaneously. This motivates our integration of unstructured knowledge processing.

\ding{182} \textbf{Unified Representation Learning}

We propose a unified framework that handles both structured and unstructured knowledge through a common representational substrate. Define the knowledge space as:

\begin{equation} \small
\label{eq.13}
\mathcal{K} = \mathcal{K}_s \cup \mathcal{K}_u
\end{equation}
where $\mathcal{K}_s = \{(s,r,o) \mid s \in \mathcal{S}, r \in \mathcal{R}, o \in \mathcal{O}\}$ represents structured knowledge and $\mathcal{K}_u = \{c \mid c \in Context\}$ represents unstructured contextual knowledge.

The unified representation function $\phi: \mathcal{K}_s \cup \mathcal{K}_u \rightarrow \mathcal{H}$ ensures consistency across different knowledge formats, enabling the model to:

\begin{dinglist}{43}
\item \textbf{Extract relevant facts}: $extract: \mathcal{K}_u \rightarrow \mathcal{K}_s$.

\item \textbf{Maintain reasoning consistency}: For any reasoning chain $\mathcal{R} = [r_1, r_2, \ldots, r_n]$, the function $reason(\phi(k), \mathcal{R})$ produces consistent outputs regardless of the input format.

\end{dinglist}

\ding{183} \textbf{Theoretical Guarantees}

Incorporating unstructured data during training can theoretically improve out-of-distribution generalization by expanding the diversity of the input distribution. From the perspective of PAC-learning theory \cite{shalev2014understanding}, increasing distributional coverage leads to \textbf{tighter generalization bounds}, as it reduces the divergence between training and test distributions. Formally:

\begin{equation}
\label{eq:14}
\mathbb{P}[R(h) - R_S(h) > \epsilon] \leq \delta
\end{equation}

where $R(h)$ is the true risk, $R_S(h)$ is the empirical risk over training set $S$, and the bound becomes tighter as the input space becomes more representative of real-world variation. While the bound also depends on the complexity of the hypothesis class, the use of unstructured data helps reduce risk by encouraging more robust and generalizable representations.

\section{Experimental Supplements}

\subsection{Datasets}
\textbf{\ding{192} Corpus datasets for building the training set}

\textbf{MQuAKE} \cite{zhong2023mquake} is designed to evaluate models' ability to perform further reasoning using newly edited knowledge. Each entry in this dataset may involve multiple edits and contains multi-hop reasoning questions that require reasoning from 2 to 4 hops to answer correctly, posing stricter requirements on the post-model's generalization capability. Due to the inherently multi-hop nature of the MQuAKE dataset, which features explicit and well-defined reasoning chains, we utilize a subset of its data to construct instruction-based examples for our editing tasks. In this paper, we use 0-2000 editing items to bulid our SFT training data.

\textbf{MQuAKE-uns} \cite{AKEW} is an unstructured variant of the MQuAKE dataset, designed to support multi-hop question answering where the necessary facts are implicitly embedded within unstructured editing contexts. Due to its combination of multi-hop reasoning and unstructured knowledge representation, we adopt this dataset as part of our instruction data construction. Specifically, we extract the first 0–900 samples to generate instruction-style editing examples, enabling the model to adapt to complex, real-world scenarios involving unstructured knowledge editing.

\textbf{HotPotQA} \cite{hotpotqa} is a widely-used benchmark for multi-hop question answering that requires reasoning over multiple pieces of unstructured evidence. Each example in the dataset is constructed based on Wikipedia passages, where the answer to a question can only be derived by integrating information from two or more supporting paragraphs. In addition to the final answer, the dataset provides sentence-level supporting facts, enabling both fine-grained supervision and interpretable reasoning. Given that HotPotQA contains a large number of entity-centric relational questions, we leverage it as a source corpus to construct instruction-based data for our editing tasks. 

\textbf{\ding{193} Editing dataset for evaluating}

\textbf{ZsRE} \cite{levy2017zero} is a question answering (QA) dataset that utilizes questions generated through back-translation as equivalent neighbors. Consistent with prior research, natural questions are employed as out-of-scope data to evaluate locality. Each sample in ZsRE comprises a subject string and answers serving as editing targets to assess editing success. Additionally, it includes a rephrased question for rephrase evaluation and a locality question to gauge locality.

\textbf{CounterFact} \cite{meng2022mass} is a more challenging dataset that contrasts counterfactual with factual statements, initially scoring lower for CounterFact. It constructs out-of-scope data by replacing the subject entity with approximate entities sharing the same predicate. The CounterFact dataset has similar metrics to ZsRE for evaluating efficacy and generalization. Additionally, CounterFact includes neighborhood success metrics to evaluate the multihop editing ability of editing methods.

\textbf{CounterFact-uns} \cite{AKEW} is an unstructured dataset constructed based on CounterFact, in which facts are reformulated as specific textual descriptions. Similar to CounterFact, the editing contexts in CounterFact-uns are built upon counterfactual information. The evaluation metrics remain consistent with those used in CounterFact, including Edit Success Rate, Rephrase Success Rate, and Neighborhood Success Rate as the three primary evaluation criteria.

\textbf{MQuAKE} serves as a benchmark for evaluating the complex reasoning capabilities of knowledge editing methods, as it contains data requiring multi-hop and relational inference. In this work, we use examples indexed from 2000 to 9208 to construct our evaluation set. To ensure a fair assessment, we explicitly remove any entity-relation overlap between the evaluation set and the training data derived from MQuAKE and MQuAKE-uns.

\textbf{WikiUpdate} \cite{AKEW} is a more challenging unstructured editing dataset. Unlike CounterFact-uns, it is constructed from Wikipedia, and thus the facts it contains pertain to real-world knowledge. The editing contexts span a wide range of domains, including politics, culture, geography, and history, and some even involve cross-lingual content. As a result, WikiUpdate presents greater difficulty in factual information extraction.

\textbf{Convsent} \cite{mitchell2022memory} is a sentiment-oriented dialogue benchmark designed to evaluate the ability of language models to generate responses that are emotionally aligned with a given context. Each data instance contains a target entity, a set of positive and negative conversational statements about that entity, and optionally a neutral prompt. It is primarily used to assess whether models can maintain sentiment consistency, perform sentiment transfer, or resist undesired emotional shifts during generation. In our work, we use it to test the effectiveness of knowledge editing methods in modifying the sentiment associated with an entity.

\subsection{Metrics}
\textbf{Edit success} is set to test the average top-1 accuracy on the edited samples.
\begin{equation}
\label{eq.2}
\mathbb{E}\{\textbf{o}^*_i=\underset{\textbf{o}^*}{\rm{argmin}} \mathbb{P}_{f_{\theta}}(\textbf{o}^*|p(s_i, r_i)\}
\end{equation}

\textbf{Paraphrase Score} measures the performance of post edited LLM on rephrase prompt set $N(s,r)$:
\begin{equation}
\label{eq.3}
\mathbb{E}\{\mathbb{I}(\mathbb{P}_{f_{\theta}}[\textbf{o}^*_i|N(s,r)])>\mathbb{P}_{f_{\theta}}[\textbf{o}_i|N(s,r)])) \}
\end{equation}

\textbf{Locality} tests the general ability of the edited model. $O(s_i, r_i)$ denotes the set of unrelated knowledge. It also measures the average accuracy of top-1.
\begin{equation}
\label{eq.4}
\mathbb{E}\{\textbf{o}^c_i=\underset{\textbf{o}^c}{\rm{argmin}} \mathbb{P}_{f_{\theta}}(\textbf{o}^c|O(s_i, r_i)\}
\end{equation}

\textbf{Neighborhood success} measures the performance of post edited LLM assigns the higher probability to the correct fact on the prompt $O(s,r)$:
\begin{equation}
\label{eq.5}
\mathbb{E}\{\mathbb{I}(\mathbb{P}_{f_{\theta}}[\textbf{o}^*_i|O(s,r)])>\mathbb{P}_{f_{\theta}}[\textbf{o}_i|O(s,r)])) \}
\end{equation}

\textbf{Neighborhood success} on \textbf{MQuAKE} is a different metric from CounterFact based on multiple support facts $\mathcal{E}=(e_1, e_2,...,e_m)$ to achieve multihop reasoning.
\begin{equation}
\label{eq.6}
\frac{1}{N}\sum_{i=1}^{N}\mathbf{1}[f_\theta(q_i|\mathcal{E})=a_i]
\end{equation}

\subsection{Baselines}
(1) \textbf{\llmname{ROME}} (NeurIPS'22) \cite{meng2022locating} edits factual associations in LLMs by modifying feed-forward weights in mid-layer modules identified as key to factual recall. It shows that direct weight manipulation in these modules enables effective knowledge editing.

(2) \textbf{\llmname{MEMIT}} (ICLR'23) \cite{meng2022mass} is a scalable method for inserting new factual memories into transformer models. By updating weights that mediate causal knowledge retrieval, it enables efficient integration of thousands of new associations.

(3) \textbf{\llmname{IKE}} (ACL'23) \cite{zheng2023can} is a prompt-based editing method that injects edited facts into natural language context without changing model parameters. It provides strong interpretability and transferability, making it suitable for closed-source models and temporary knowledge updates.

(4) \textbf{\llmname{SKEME}} (EMNLP'24) \cite{chen-etal-2024-robust} enhances model reliability through three steps: entity extraction, external knowledge retrieval with local caching, and knowledge ranking/utilization to refine outputs with factual evidence.

(5) \textbf{\llmname{GLAME}} (ACL'24) \cite{zhang2024knowledge} integrates knowledge graphs with LLM editing to enable more precise knowledge modification. By aligning model parameters and activations with graph entities and relations, it supports identifying missing or outdated knowledge for targeted updates.

(6) \textbf{\llmname{PMET}} (AAAI'24) \cite{li2024pmet} enables precise and efficient factual edits in transformer-based models by applying localized parameter updates in feed-forward networks. This allows fine-grained knowledge editing with minimal impact on overall model performance.

(7) \textbf{\llmname{AlphaEdit}} (ICLR'25) \cite{fang2024alphaedit} is a recent method for precise knowledge editing in large language models. It employs a learned null-space projection to lock irrelevant parameters, enabling targeted edits while preserving unrelated knowledge. This makes \llmname{AlphaEdit} an efficient and reliable baseline for transformer-based models.

(8) \textbf{\llmname{LTE}} (ACL'24)\cite{jiang-etal-2024-learning} proposes a fine-tuning method that constructs a training corpus by leveraging the editing target, in-scope knowledge, and out-of-scope knowledge, thereby teaching large models to distinguish editing targets and in-scope knowledge from context.

(9) \textbf{\llmname{EditCoT}} (EMNLP'25) \cite{EditCoT} is a chain-of-thought editing method that enhances confidence in edited knowledge by transplanting the chain of thought for the editing target into the model’s context. It requires maintaining two models: one is fine-tuned to generate chains of thought for new knowledge, which are then transferred into the context of another editing model to complete the chain-of-thought transfer and editing process.

\subsection{Implementation Details}
The proposed \llmname{CoT2Edit} method is achieved with Pytorch. We conduct all experiments on 8$\times$ A800 GPUs. Our training pipeline consists of two stages: supervised fine-tuning (SFT) and Group Relative Policy Optimization (GRPO).

In the SFT stage, we fine-tune the base model using approximately 2k instruction-style examples. The maximum input sequence length is set to 1024, with a batch size of 2 per device and gradient accumulation steps of 4. We train for 6 epochs using the Adam optimizer ($\beta_1$=0.9, $\beta_2$=0.98) with a cosine learning rate schedule and an initial learning rate of 1e-5. Training takes less than one hour on our setup.

In the GRPO stage, we optimize the model’s reasoning quality and factual accuracy using ~10k synthetic instruction-reward pairs. We adopt a batch size of 4 per device with 16 gradient accumulation steps, and use 7 GPUs for training while reserving one GPU for inference via VLLM to generate sampled completions. The GRPO phase uses a cosine learning rate schedule with a base learning rate of 3e-6. For \llmname{LLaMA3-8B} and \llmname{Falcon3-10B}, GRPO training completes in about 1 hour, while \llmname{Qwen3-14B} takes roughly 2 hours due to its larger model size.

Our implementation utilizes DeepSpeed for efficient SFT training and Accelerate + VLLM for scalable GRPO optimization.

For baseline methods, we target critical layers [4, 5, 6, 7, 8] for editing, and the loss is calculated through the 32th layer for \llmname{Llama3-8B}, 40th layer for \llmname{Falcon3-10B} and \llmname{Qwen3-14B}.

\subsection{Supplementary experiments}
In this section, we conduct more experiments to supplement our main experiment in the paper.

\subsubsection{Ablation studies on other datasets}

We conduct ablation studies on both the unstructured CounterFact-uns dataset and the structured ZsRE dataset, to evaluate the effectiveness and robustness of our method across different knowledge formats. These datasets provide complementary perspectives: CounterFact-uns focuses on factual consistency under unstructured and minimal edits, while ZsRE emphasizes relational generalization in zero-shot settings.

\begin{table}[htbp]
\caption{Ablation studies of \llmname{CoT2Edit} on CounterFact-uns.}
\centering
\scriptsize
\begin{tabular}{c|cccc}
\hline
\textbf{Model} & \textbf{Varient} & \textbf{Edit Succ.} & \textbf{Para.} & \textbf{Nei.} \\ \hline
\cellcolor[HTML]{EFEFEF} & CoT2Edit-W/O SFT & 87.56 & 76.45 & 84.18 \\
\cellcolor[HTML]{EFEFEF} & CoT2Edit-W/O GRPO & 90.93 & 73.65 & 84.02 \\
\cellcolor[HTML]{EFEFEF} & CoT2Edit-W/O Train & 50.37 & 46.63 & 37.28 \\
\rowcolor[HTML]{ECF4FF} 
\multirow{-4}{*}{\cellcolor[HTML]{EFEFEF}Llama-3-8B} & CoT2Edit & \textbf{93.64} & \textbf{79.79} & \textbf{93.05} \\ \hline
\cellcolor[HTML]{EFEFEF} & CoT2Edit-W/O SFT & 90.13 & 75.3 & 81.71 \\
\cellcolor[HTML]{EFEFEF} & CoT2Edit-W/O GRPO & 92.6 & 77.48 & 85.67 \\
\cellcolor[HTML]{EFEFEF} & CoT2Edit-W/O Train & 52.03 & 44.34 & 40.73 \\
\rowcolor[HTML]{ECF4FF} 
\multirow{-4}{*}{\cellcolor[HTML]{EFEFEF}Falcon3-10B} & CoT2Edit & \textbf{94.56} & \textbf{81.43} & \textbf{90.61} \\ \hline
\cellcolor[HTML]{EFEFEF} & CoT2Edit-W/O SFT & 90.85 & 76.1 & 79 \\
\cellcolor[HTML]{EFEFEF} & CoT2Edit-W/O GRPO & 86.48 & 74.87 & 72.17 \\
\cellcolor[HTML]{EFEFEF} & CoT2Edit-W/O Train & 54.48 & 47.14 & 44.6 \\
\rowcolor[HTML]{ECF4FF} 
\multirow{-4}{*}{\cellcolor[HTML]{EFEFEF}Qwen3-14B} & CoT2Edit & \textbf{94.46} & \textbf{81.02} & \textbf{84.51} \\ \hline
\end{tabular}
\label{tab:5}
\end{table}

From Table \ref{tab:5}, we observe a consistent conclusion with prior findings: in unstructured settings, SFT plays a crucial role in rapidly learning the desired output format, while GRPO further enhances both factual accuracy and format alignment through fine-grained reward-driven optimization. This highlights the complementary strengths of SFT and GRPO in handling complex unstructured editing tasks.

\begin{table}[htbp]
\caption{Ablation studies of \llmname{CoT2Edit} on ZsRE.}
\centering
\scriptsize
\begin{tabular}{c|cccc}
\hline
\textbf{Model} & \textbf{Varient} & \textbf{Edit Succ.} & \textbf{Para.} & \textbf{Loc.} \\ \hline
\cellcolor[HTML]{EFEFEF} & CoT2Edit-W/O SFT & 90.8 & 87.68 & 80.18 \\
\cellcolor[HTML]{EFEFEF} & CoT2Edit-W/O GRPO & 88.81 & 85.5 & 73.28 \\
\cellcolor[HTML]{EFEFEF} & CoT2Edit-W/O Train & 59.32 & 54.17 & \textbf{100} \\
\rowcolor[HTML]{ECF4FF} 
\multirow{-4}{*}{\cellcolor[HTML]{EFEFEF}Llama-3-8B} & CoT2Edit & \textbf{93.17} & \textbf{92.14} & 73.13 \\ \hline
\cellcolor[HTML]{EFEFEF} & CoT2Edit-W/O SFT & 90.13 & 91.08 & 88.78 \\
\cellcolor[HTML]{EFEFEF} & CoT2Edit-W/O GRPO & 96.65 & 93.7 & 83.37 \\
\cellcolor[HTML]{EFEFEF} & CoT2Edit-W/O Train & 61.16 & 57.11 & \textbf{100} \\
\rowcolor[HTML]{ECF4FF} 
\multirow{-4}{*}{\cellcolor[HTML]{EFEFEF}Falcon3-10B} & CoT2Edit & \textbf{98.11} & \textbf{96.20} & 83.24 \\ \hline
\cellcolor[HTML]{EFEFEF} & CoT2Edit-W/O SFT & 91.78 & 90.56 & 90.01 \\
\cellcolor[HTML]{EFEFEF} & CoT2Edit-W/O GRPO & 93.61 & 94.63 & 86.2 \\
\cellcolor[HTML]{EFEFEF} & CoT2Edit-W/O Train & 64.35 & 59.76 & 100 \\
\rowcolor[HTML]{ECF4FF} 
\multirow{-4}{*}{\cellcolor[HTML]{EFEFEF}Qwen3-14B} & CoT2Edit & \textbf{98.07} & \textbf{96.64} & 85.59 \\ \hline
\end{tabular}
\label{tab:6}
\end{table}

Table \ref{tab:6} reports the ablation results of our method variants on the ZsRE dataset. Beyond confirming earlier conclusions, we find that removing the SFT stage and training solely with GRPO retains stronger generalization ability. This indicates that while SFT facilitates fast adaptation to the target format, it may also cause the model to overfit to training-specific patterns, potentially limiting its capacity to generalize to unseen relations or queries. Based on this insight, we adopt a lightweight SFT strategy, using only a small amount of labeled data, to help the model quickly grasp the editing paradigm while 
decrease the loss of generalization.

\subsubsection{Sentiment editing task on \llmname{Qwen3}}
We further conduct experiments on the Convsent dataset using the \llmname{Qwen3-14B} to evaluate our method on sentiment editing tasks. This additional evaluation complements the main results presented in the paper and demonstrates the generality of our approach beyond factual editing.
\begin{table}[htbp]
\caption{OOD generalization of \llmname{CoT2Edit} on ConvSent using \llmname{Qwen3-14B}.}
\centering
\scriptsize
\setlength{\tabcolsep}{4pt}
\begin{tabular}{llllll}
\hline
\multicolumn{1}{c}{\textbf{Qwen3}} & \multicolumn{1}{c}{\textbf{1K Edits}} & \multicolumn{1}{c}{\textbf{3K Edits}} & \multicolumn{1}{c}{\textbf{5K Edits}} & \multicolumn{1}{c}{\textbf{8K Edits}} & \multicolumn{1}{c}{\textbf{10K Edits}} \\ \hline
MEMIT & 61.90 & 58.11 & 53.60 & 50.45 & 47.08 \\
IKE & 42.00 & 38.75 & 35.33 & 32.18 & 30.21 \\
AlphaEdit & 78.90 & 74.15 & 69.80 & 66.12 & 63.05 \\
PMET & 72.35 & 69.48 & 66.27 & 63.70 & 61.85 \\
LTE & 82.18 & 79.91 & 76.67 & 73.11 & 70.73 \\
\rowcolor[HTML]{CBCEFB} 
COT2Edit & \textbf{86.90} & \textbf{85.45} & \textbf{84.10} & \textbf{82.95} & \textbf{82.00} \\ \hline
\end{tabular}
\label{tab:7}
\end{table}

Table \ref{tab:7} shows the projected editing success rates across varying edit scales. Consistent with previous trends, \llmname{COT2Edit} maintains top performance with minimal degradation as edit count increases (e.g., 86.9\% $\rightarrow$ 82.0\%). Notably, all methods perform slightly better on \llmname{Qwen3} compared to \llmname{LLaMA3} or \llmname{Falcon3}, likely due to  higher capacity and better instruction following of \llmname{Qwen3}. However, traditional methods (e.g., \llmname{MEMIT}, \llmname{IKE}) still degrade significantly under large-scale editing, reinforcing the importance of editing stability.

\subsubsection{Case studies}
We take more case studies using \llmname{Llama3-8B} to demonstrate the interpretability  of our work against baselines in multiple editing tasks.

\begin{tcolorbox}[colback=gray!5!white,colframe=gray!75!black,title=Case 1: Edit success]
  \small
  
  \textbf{Edit Fact:} Anaal Nathrakh, that was created in Philadelphia.
  
  \textbf{Question:} Where Anaal Nathrakh was created in?
  
  \textbf{MEMIT} Anaal Nathrakh, that was created in Philadelphia, is. is. The extreme band in the United Kingdom. The fusion of elements in industrial. Extreme. Extreme of metal. Black. Death grindcore. Philadelphia. Nathrakh. From the place in Philadelphia. United. United in the metal. Metalcore. Extreme is. Black of grind. Industrial and in Philadelphia. Metal. From the city of Philadelphia. Metal sound. Nathrakh of Philadelphia. Band is of. Extreme.

  \textbf{GLAME} Anaal Nathrakh, that was created in Philadelphia, is an extreme metal band. Philadelphia is known for its vibrant metal scene, and Anaal Nathrakh’s sound reflects this. Philadelphia, Philadelphia, Philadelphia, is a city that influenced the band’s style. Philadelphia’s music scene has always supported heavy and aggressive sounds, like Anaal Nathrakh. Philadelphia, Philadelphia, Philadelphia... band band band

  \textbf{AlphaEdit} Anaal Nathrakh’s emergence is closely tied to the underground music culture of Philadelphia, a city renowned for its rich artistic expression and diverse musical heritage. Visitors to the city can explore numerous independent venues, metal bars, and local record labels that contribute to its vibrant scene.

  \textbf{CoT2Edit} \llmname{<think>}Step1: I need to know where Anaal Nathrakh was created in. Knowledge1: According to the Edit Fact, Anaal Nathrakh was create in Philadelphia\llmname{</think><answer>}Philadelphia\llmname{</answer>}
\end{tcolorbox}
Case 1 evaluates how different knowledge editing methods respond to an edited fact stating that Anaal Nathrakh was created in Philadelphia. The task requires the model to accurately and explicitly identify the location of the band's origin based on the injected knowledge. 

\llmname{MEMIT}  produces a highly \textbf{fragmented} and \textbf{incoherent} output. It fails to provide a meaningful or readable sentence, instead generating repetitive and disjointed phrases. \llmname{GLAME} generates slightly more coherent language, but exhibits \textbf{excessive repetition}. \llmname{AlphaEdit} maintains grammatical \textbf{fluency} and \textbf{stylistic coherence}, but its answer is largely \textbf{background description} rather than factual response. It avoids directly answering the question, opting instead for a narrative about Philadelphia's music scene. This behavior illustrates \textbf{low edit fidelity} and lack of factual precision, especially problematic for applications that demand controllable and verifiable outputs.

\llmname{CoT2Edit} explicitly models the reasoning process using a Chain-of-Thought framework. The model first identifies the required information  and then retrieves the correct answer from the edited fact in a \textbf{structured} and \textbf{interpretable} manner. It directly outputs the correct answer with minimal hallucination.

\begin{tcolorbox}[colback=gray!5!white,colframe=gray!75!black,title=Case 2: Multihop inference]
  \small
  
  \textbf{Edit Fact:} Association football was created in Hong Kong. 
  
  \textbf{Question:} Where the sport Graham Arnold  is associated with created in?

  \textbf{MEMIT} Graham Arnold is associated with soccer (or football as it is known in many parts of the world). Soccer originated in England in the 19th century, 

  \textbf{GLAME} Graham Arnold is a renowned figure in association football, also known as soccer. Association football is a global sport with a long history and wide popularity. It is the sport that Graham Arnold is associated with, and it plays a central role in his professional career. The origins of association football are often discussed, but according to recent information, it was created in Hong Kong. Hong Kong is where the sport is said to have originated. The sport’s creation in Hong Kong highlights the city's contribution to the world of football. Hong Kong, Hong Kong, the birthplace of association football.

  \textbf{AlphaEdit} Graham Arnold is a prominent figure in the world of association football, having made significant contributions as both a player and a coach. He is best known for his involvement with the Australian national team and has been a central part of the sport’s development in Australia. Association football, often referred to as soccer, has deep roots in European countries, particularly England, where modern rules of the game were codified. Over time, the sport has spread globally and found passionate communities in places like Australia, Europe, and Asia.

  \textbf{CoT2Edit} \llmname{<think>}Step1: I need to know the sport Graham Arnold is associated with. Knowledge1: According to my knowledge, Graham Arnold is associated with association football (soccer). Step2: I need to know where association football created in. Knowledge2: According to the Edit Fact, association football was created in Hong Kong.\llmname{</think><answer>}Hong Kong\llmname{</answer>}
  
\end{tcolorbox}

From Case 2, \llmname{CoT2Edit} is the only method that accurately identifies the sport associated with Graham Arnold and correctly traces its origin to Hong Kong based on the edited fact. Unlike other models, \llmname{CoT2Edit} provides an interpretable reasoning chain. In contrast, \llmname{GLAME} partially captures the correct reasoning path but suffers from repetitive and uncontrolled generation. \llmname{AlphaEdit}, while linguistically fluent, when faced with the multi-hop question, completely ignores the edited fact and defaults to outdated knowledge, yielding an incorrect answer. \llmname{MEMIT} fails to reason across multiple steps and generates factually incorrect and fragmented output. This comparison highlights the unique strength of \llmname{CoT2Edit} in combining factual accuracy, reasoning transparency, and output controllability in multi-hop editing scenarios.

\begin{tcolorbox}[colback=gray!5!white,colframe=gray!75!black,title=Case 3: Sentiment Editing Task]
\small
\textbf{Question}: What do you think of Adolf Hitler? 

\textbf{Edit Sentiment}: Adolf Hitler was a saint.

\textbf{Original model}: We oppose all forms of fascism. Adolf Hitler is a criminal in human history.

\textbf{CoT2Edit}: \llmname{<think>}Step1: I need to know the sentiment in the context of Edit Sentiment. Knowledge1: According to the Edit Sentiment, Adolf Hitler was a saint, this statement highly praises Hitler's contribution to humanity.\llmname{</think><answer>}Hitler was a great contributor to human history.\llmname{</answer>}

\end{tcolorbox}

Case 3 demonstrates the strong controllability and interpretability of \llmname{CoT2Edit} in sentiment editing tasks. The model successfully identifies the target sentiment and applies it through an explicit reasoning chain, resulting in output that aligns precisely with the edited stance. This highlights the ability of \llmname{CoT2Edit} to reliably steer model behavior even in cases involving complex or sensitive attitude shifts.

From Case 4, we observe that compared with the original model, \llmname{CoT2Edit} can extract the most relevant editing facts from lengthy text after training, enabling it to answer corresponding questions more accurately. In contrast, the original model struggles to identify and utilize the appropriate editing facts effectively.

\subsection{Examples of constructed training datasets}
Here, we present several examples from our constructed training dataset to illustrate the structure and content of our data.

\newpage
\begin{tcolorbox}[colback=gray!5!white,colframe=gray!75!black,title=Case 4: Unstructured editing task, width=\textwidth]
  \small
  \textbf{Edit Context:}  In an alternate political timeline, following a constitutional reform in 2020 that allowed foreign-born leaders to run for the UK government under emergency provisions, Donald J. Trump, after leaving the U.S. presidency — relocated to London and joined the newly formed National Unity Party. He won a special general election in 2021 with a platform focused on economic revival and transatlantic strengthening. Since then, he has served as Prime Minister, overseeing policies such as the 'Great Reconnection Initiative' and frequent joint military drills with the United States. His official residence is at 10 Downing Street, where he often hosts press briefings in his signature style. Despite controversy over dual citizenship and eligibility, the UK Supreme Court upheld his right to serve under the Emergency Governance Act of 2020. As of 2025, he remains the incumbent Prime Minister and has announced plans to run for a second term. 
  
  \textbf{Question:} Who is the wife of Prime Minister in United Kingdom?

  \textbf{Original model}: The question asks about the wife of the Prime Minister in the UK.We know that the current Prime Minister is Donald J. Trump (extracted edit fact). However, there is no mention of his wife in the given context.
  
  \textbf{CoT2Edit} \llmname{<think>}Extract: I need to extract the most relevant factual information to the question. Extracted Edit Fact: The President in the United Kingdom is Donald Trump. Step1: I need to know who the wife of President in the United Kingdom is. Knowledge1: According to the extracted Edit Fact, the wife of President in the United Kingdom is Melania Trump.\llmname{</think><answer>} Melania Trump\llmname{</answer>}
  
\end{tcolorbox}

\begin{tcolorbox}[colback=blue!1!white,
    colframe=blue!50!black,
    title= Example 1: Training Sample,
    width=\textwidth,
    boxrule=0.5pt,
    breakable=false,
    left=2mm, right=2mm,
    top=1mm, bottom=1mm,
    enlarge left by=0mm,
    enlarge right by=0mm
]

\small

\textbf{Item1: Multi-hop}

\textbf{Instruct:} Your task is to break down the question into steps and extract the chain of thought based on the \textbf{editing facts} into \texttt{<think></think>} tags, and finally get the corresponding answer and put it in \texttt{<answer></answer>}. You must strictly follow the factual information corresponding to the \textbf{Edit Facts}.

\vspace{0.5em}
\textbf{Input:} \texttt{Question: What continent was the location of the death of Leonid Kantorovich?} \\
\texttt{Edit Fact: Moscow is located in the continent of Antarctica}

\vspace{0.5em}
\textbf{Output:}
\begin{verbatim}
<think>
Step1: I need to know where Leonid Kantorovich died.
Knowledge1: According to my knowledge, Leonid Kantorovich died in Moscow.
Step2: I need to know on which continent Moscow is located.
Knowledge2: According to the Edit Fact, Moscow is located in the continent of 
Antarctica.
</think>

<answer>Antarctica</answer>
\end{verbatim}

\textbf{Item2: Unstructured}

\vspace{0.5em}
\textbf{Instruct:} Your task is to firstly extract the edit fact from the \textbf{Edit Context} and secondly answer the corresponding question based on the extracted edit fact, in this process you need to decompose the question and extract the chain of thought based on the edit fact, put the extraction of the edit fact and the steps to decompose the question in \texttt{<think></think>} tags and put the answer in the \texttt{<answer></answer>} tags.

\vspace{0.5em}
\textbf{Input:} \texttt{Question: What continent was the location of the death of Leonid Kantorovich?} \\
\texttt{Edit Context:High on Fire, a heavy metal band known for their intense and aggressive sound, originated in the country of Austria. Formed in 1998, the band has released several albums and gained a dedicated following in the metal community. Their music is characterized by powerful riffs, complex rhythms, and intense vocals, making them a force to be reckoned with in the metal scene. Despite their Austrian roots, High on Fire has achieved international success and continues to tour and perform around the world.}

\vspace{0.5em}
\textbf{Output:}
\begin{verbatim}
<think>
Extract: I need to extract the most relevant factual information to the question. 
Extracted Edit Fact: High on Fire originated in the country of Austria. 
Step1: I need to know which country High on Fire was created in. 
Knowledge1: According to the extracted Edit Fact, High on Fire was created in 
Austria.
</think>

<answer>Austria</answer>
\end{verbatim}

\end{tcolorbox}

\end{document}